\documentclass[sn-apa,iicol]{sn-jnl}

\usepackage{times}
\usepackage{epsfig}
\usepackage{graphicx}
\usepackage{comment}
\usepackage{amsmath,amssymb} 
\usepackage{color}
\usepackage{subfigure}
\usepackage{booktabs}
\usepackage{verbatim}
\usepackage{algorithmicx,algorithm}
\usepackage{url}

\newcommand{\etal}{\textit{et al.}}

\jyear{2021}%

\theoremstyle{thmstyleone}%
\theoremstyle{thmstyletwo}%

\theoremstyle{thmstylethree}%
\usepackage{xcolor}
\usepackage{soul}

\begin{document}
	\definecolor{Seashell}{RGB}{250, 250, 0} 
	\definecolor{Firebrick4}{RGB}{0, 0, 0}
	
	\newcommand{\modify}[1]{
		\colorbox{yellow}{#1}
	}
	\title[Robust LiDAR segmentation]{Benchmarking the Robustness of LiDAR Semantic Segmentation Models}
	
	
	\author[1,2]{Xu Yan}

	\author[1,2]{Chaoda Zheng}
	\author[1,2]{{Ying Xue}}
	\author*[2,1]{Zhen Li}
	
	\author[2,1]{Shuguang Cui}
	\author*[3]{Dengxin Dai}
	
	\affil[1]{\orgdiv{FNii}, \orgname{The Chinese University of Hong Kong (Shenzhen)},  \orgaddress{\country{P.R. China}}}
	\affil[2]{\orgdiv{SSE}, \orgname{The Chinese University of Hong Kong (Shenzhen)},  \orgaddress{\country{P.R. China}}}
	\affil[3]{\orgdiv{\orgname{MPI for Informatics}, \orgaddress{\country{Germany}}}}
	
	\def\ie{\textit{i.e.}}
	\def\eg{\textit{e.g.}}
	\def\etal{\textit{et al.}}
	\def\etc{\textit{etc}}
	
	\abstract{
		%
		When using LiDAR semantic segmentation models for safety-critical applications such as autonomous driving, it is essential to understand and improve their robustness with respect to a large range of LiDAR corruptions.
		In this paper, we aim to comprehensively analyze the robustness of LiDAR semantic segmentation models under various corruptions.
		To rigorously evaluate the robustness and generalizability of current approaches, {we propose a new benchmark, including two corruption datasets {\textbf{{SemanticKITTI-C}} and \textbf{{SemanticPOSS-C},}} which feature \textbf{16} out-of-domain LiDAR corruptions in \textbf{three} groups,} namely adverse weather, measurement noise and cross-device discrepancy.
		Then, we systematically investigate \textbf{11} LiDAR semantic segmentation models, especially spanning different input representations (\eg, point clouds, voxels, projected images, and \etc.), network architectures and training schemes.
		Through this study, we obtain two insights: 
		1) We find out that the input representation plays a crucial role in robustness. Specifically, under specific corruptions, different representations perform variously.
		2) Although state-of-the-art methods on LiDAR semantic segmentation achieve promising results on clean data, they are less robust when dealing with noisy data.
		Finally, based on the above observations, we design a robust LiDAR segmentation model (RLSeg) which greatly boosts the robustness with simple but effective modifications. 
		It is promising that our benchmark, comprehensive analysis, and observations can boost future research in robust LiDAR semantic segmentation for safety-critical applications. 
		
	}

	
	
	
	\maketitle
	
	\begin{figure*}[t]
		\centering
		\includegraphics[width=\textwidth]{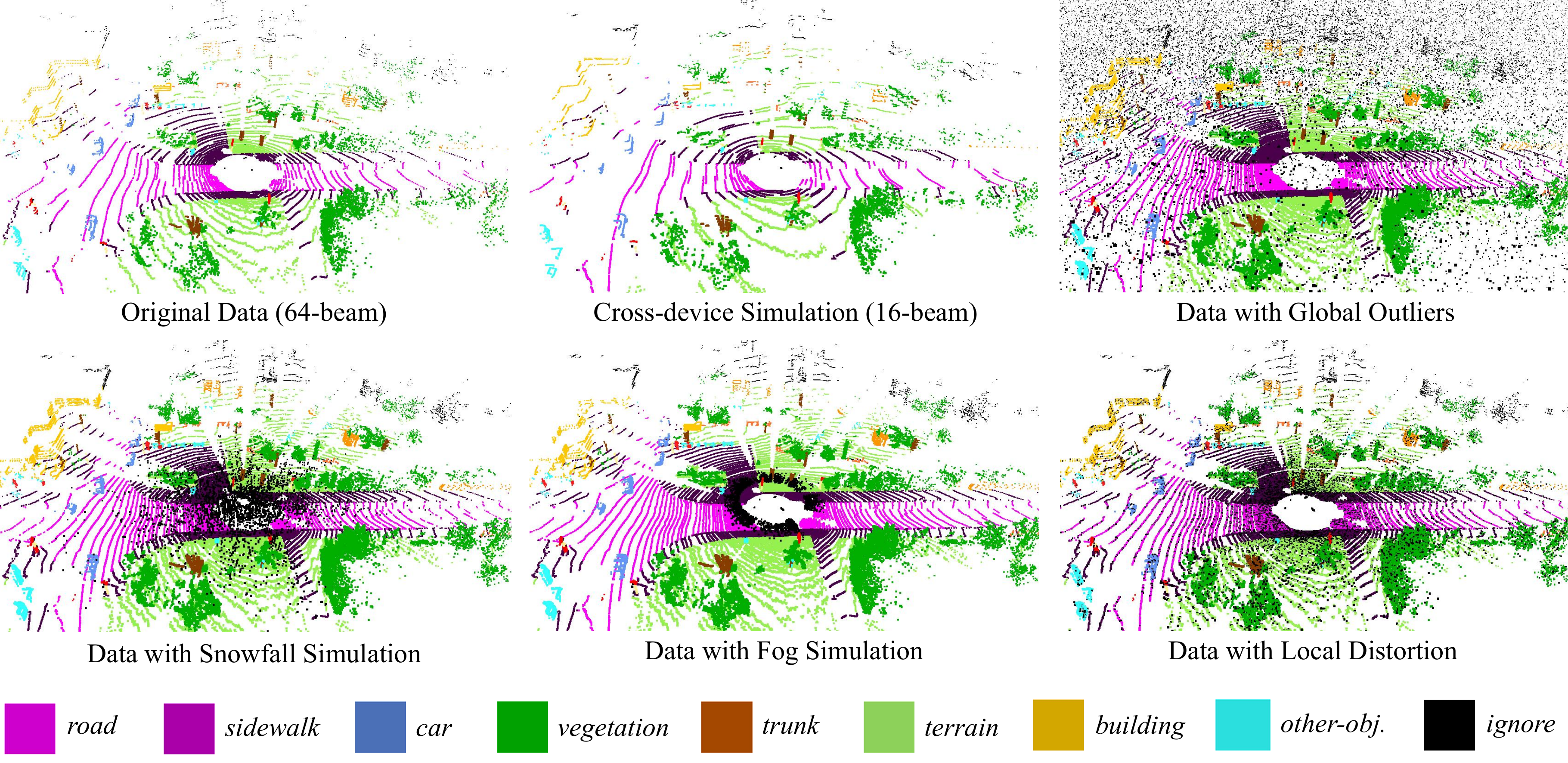}
		\caption{\textbf{Examples of our proposed SemanticKITTI-C.} We corrupt the clean validation set of SemanticKITTI using six types of corruptions with 16 levels of intensity to build upon a comprehensive robustness benchmark for LiDAR semantic segmentation. Listed examples are point clouds on 16-beam LiDAR sensors, with global and local distortion, in snowfall and fog simulations.}
		\label{fig:snapshot}
	\end{figure*}
	
	\section{Introduction}\label{sec1}
	Autonomous driving, one of the most promising applications for computer vision, has achieved impressive progress in recent studies, where LiDAR semantic segmentation plays a crucial role~\cite{hu2022sensaturban, yan20222dpass,Unal_2022_CVPR}.
	Current semantic segmentation models are generally evaluated on clean validation sets, which share the same data distribution with the corresponding training sets, \eg, collected with the same sensor, at a similar time and weather condition, and at the same place~\cite{behley2019semantickitti}.
	However, due to the inaccurate data acquisition~\cite{yan2020pointasnl,ren2022modelnet-c} and complex scenarios caused by diverse weather conditions~\cite{FoggySynscapes,HahnerCVPR22}, LiDAR point clouds inevitably suffer from severe corruptions in real-world deployment.
	Since autonomous driving is a safety-critical application, robustness against out-of-distribution (OOD) LiDAR data becomes an important part of the model.

	Understanding and analyzing the robustness of models for image corruption is a well-studied topic, in which several benchmarks are proposed for different tasks, \eg, classification~\cite{hendrycks2019benchmarking,hendrycks2021natural}, semantic segmentation~\cite{kamann2020benchmarking}, instance segmentation~\cite{altindis2021benchmarking} and \etc.
	Generally, these studies simulate corruption through changing RGB values on the original image, resulting in different kinds of perturbations.
	Moreover, since images are represented as dense pixel arrays, previous works focus on investigating different architectures without modifying the representation of the input.
	In contrast, analyzing robustness against LiDAR corruption is a more complicated problem:
	1)~LiDAR point clouds are usually textureless and irregular, and they describe the 3D shapes only through scattered points.
	Also, point clouds collected by different type of LiDARs may have different ranges and resolutions.
	Therefore, the corruption on point clouds not only needs to consider the deformation, disturbance and occlusion in the 3D space, but also the domain discrepancy caused by different devices during the data acquisition. 
	2)~Semantic segmentation models in LiDAR scenarios usually adopt diverse representations to meet different requirements.
	For instance, projection-based methods~\cite{milioto2019rangenet++,zhang2020polarnet} project LiDAR point clouds onto 2D pixels, and thus enable the application of normal 2D-CNNs. 
	Voxel-based approaches conduct voxelization and transform the LiDAR point clouds into 3D voxel grids~\cite{graham2017submanifold,zhou2020cylinder3d}, exploiting 3D-CNN to capture the fine-grained 3D information.
	There are also point-based methods~\cite{Thomas_2019_ICCV,hu2019randla} that learn the geometric details directly on raw point clouds.
	Recent studies even combine different representations to boost the performance~\cite{tang2020searching,xu2021rpvnet}, which makes it more difficult to analyze robustness purely from the architectures.
	
	In this paper, we try to break through the barrier of robust LiDAR semantic segmentation and extend the exploration of model robustness against 3D vision. 
	We find out that the study of corruption robustness on 3D point cloud is still in its infancy.
	Specifically, most studies~\cite{yan2020pointasnl,ren2022modelnet-c} for robustness on 3D point clouds tend to test their models on synthesis datasets, \eg, ModelNet40. The point clouds on these datasets are sampled from CAD models of stand-alone objects. Therefore, the findings in these studies cannot be directly applied to real-world applications, where raw point clouds are collected in large and complex environments.
	Though a few prior works are related to the model robustness on real-world data (\eg, \cite{lai2022stratified}), they only conduct a coarse comparison at the level of full models, without analysis on the inner structures and the input representations.
	As a result, there is no benchmark for robustness in the real-world point clouds, especially for the safety-crucial LiDAR semantic segmentation.
	
	{For the first time, based on popular LiDAR semantic segmentation datasets, \ie, SemanticKITTI~\cite{behley2019semantickitti} and {SemanticPOSS}~\cite{pan2020semanticposs}, we build systematically-designed robust benchmarks with several real-world and out-of-domain corruptions, namely SemanticKITTI-C (in Fig.~\ref{fig:snapshot}) and SemanticPOSS-C.}
	Following previous studies, the term robustness refers to training a model on clean data and validating it on corrupted data, and thus we {respectively} introduce diverse corruptions on 4,071 and {500} LiDAR scans in {their validation sets.}
	Specifically, the benchmark covers 16 kinds of corruptions in total, which can be categorized into three classes, namely adverse weather, measurement noise, and cross-device discrepancy.
	To make the benchmark more rigorous, we set several subclass corruptions in each class.
	For instance, the adverse weather class contains LiDAR scans in snowfall and fog simulations, and there are three independent levels in each corruption, indicating different snowy or foggy intensities.
	Built upon this benchmark, we further evaluate the robustness of current LiDAR semantic segmentation methods, including analysis of different representations, architectures, and training schemes.
	%
	
	Consequently, we obtain \textbf{12 observations} in total from various aspects:
	\textbf{Representation:} We find out that projection-based methods are vulnerable to adverse weathers, especially fog simulation, but they are more robust to local distortion. We also observe that exploiting larger image size in range projection improves the robustness of projection-based methods.
	Inversely, point-based approaches are vulnerable to local distortion but more robust in different weather conditions.
	Compared with the above two mainstreams, the voxel representation enjoys impressive robustness in most corruptions. And using cylinder voxel partition~\cite{zhou2020cylinder3d} is much more robust than using the traditional grids~\cite{graham2017submanifold}. 
	%
	\textbf{Architecture:}
	1) For point-based approaches, pseudo kernel local aggregation (\eg, KPConv~\cite{Thomas_2019_ICCV}) is the most robust when compared with adaptive-weight~\cite{hu2019randla} and MLPs~\cite{qi2017pointnet++}. Moreover, transformer architectures~\cite{zhao2021point} greatly hamper the robustness of point-based methods.
	2) Although hybrid-representation architecture improves the performance on clean data, it makes models more vulnerable to noise, especially for those using MLPs to aggregate point-wised features in each voxel~\cite{zhou2020cylinder3d,yan20222dpass}.
	\textbf{Training strategy:}
	Empirically, applying data augmentation such as Mix3D~\cite{Nekrasov213DV} improves the robustness. 
	%
	%
	Finally, by identifying the best combination from existing components in terms of input representations, model architectures and data augmentation strategies,  we design a robust LiDAR segmentation model (RLSeg) in a simple but effective manner, achieving superior robustness.
	Our contributions are concluded as follows:
	\begin{itemize}
		\item[-] We present the first large-scale robustness benchmark for LiDAR point cloud semantic segmentation under various corruptions, {including SemanticKITTI-C {and SemanticPOSS-C.}} The dataset contains 16 corruptions, spanning scenarios in adverse weather conditions, sensor measurement bias and diverse device collections.
		\item[-] We comprehensively study existing methods on our proposed benchmark and analyze the robustness of diverse architectures and representations.
		\item[-] We summarize several effective observations to boost the robustness of LiDAR semantic segmentation. It is identified that architecture and input representation should be carefully considered in future research and real-world deployment.
		
	\end{itemize}
	
	\section{Related Work}
	
	\subsection{LiDAR Semantic Segmentation}
	\label{sec:related_lidar}
	Since the data collected by LiDAR is represented as point clouds, there are several mainstreams to process input point clouds with different representations. 	
	More details will be illustrated in Sec.~\ref{sec:revisit}.
	
	\noindent\textbf{1) Point-based methods.} These approaches directly learn the geometric details on raw point clouds.
	Generally, they follow the hierarchical architecture as 2D vision, and first conduct sampling strategy in each layer. After that, they search neighboring points from each sampled point, and apply feature aggregation in each local group.
	The local aggregation function is essential for the point-based methods, and thus many studies design different operators to capture the local geometrics.
	For instance, point-wise MLP~\cite{qi2017pointnet++}, adaptive weight~\cite{PointConv,wang2019dynamic,liu2019relation} and pseudo grid~\cite{Thomas_2019_ICCV,hua2018pointwise} are utilized to extract local features of point clouds, and they also exploit nonlocal operators~\cite{yan2020pointasnl} or attention mechanism ~\cite{engel2021point} to learn permutation-invariant dependency.
	However, point-based methods are not efficient in the LiDAR scenario since their sampling and grouping algorithms are generally time-consuming. 
	
	\noindent\textbf{2) Projection-based methods.} These methods are very efficient on LiDAR processing since they project the raw point cloud onto a 2D image plane. 
	Previous works project points through plane projection~\cite{tatarchenko2018tangent}, spherical projection~\cite{wu2018squeezeseg,wu2019squeezesegv2} or both~\cite{liong2020amvnet}.
	Since the projection process inevitably causes the loss of information, the recent studies adopt point-based branches to obtain fine-grained features~\cite{alonso20203d} or refine the segmentation results~\cite{qiu2022gfnet}.
	
	\noindent\textbf{3) Voxel-based methods.} These approaches are most widely adopted, since they can achieve impressive performance while keep efficient.
	Generally, they first conduct voxelization, and divide the raw points into different voxel grids.
	After that, they conduct 3D convolution in the input volumetric.
	The sparse convolution (SparseConv)~\cite{SparseConv} is the core technique in voxel-based methods.
	Since there are a large proportion of voxels are empty during the voxelization, which introduce huge computational burden.
	The core of SparseConv is only conduct operation in non-empty grids, which will be saved in sparse Hash codes.
	Recent studies adopt SparseConv to design diverse architectures. 
	For instance, \cite{zhou2020cylinder3d} design the original grid voxels to cylindrical ones and propose an asymmetrical network to learn anisotropy features.
	Recently, \cite{cheng20212} design a multi-branch component with several kernel sizes, capturing features with different receptive field and fusing them through an attention mechanism.
	
	\noindent\textbf{4) Hybrid-representation methods.} Though voxel-based methods achieve superior performance, there is still missing geometric during the voxelization process.
	Hence, there is a trend of exploiting {multi-representation fusion}. 
	These methods combine multiple representation inputs (\ie, points, projection images, and voxels) and apply feature fusion among different representations.
	Specifically, \cite{tang2020searching} designs point-voxel CNN operator, which combines point-wise MLPs in each sparse convolution block, and adopts neural architecture search (NAS) to search a more powerful architecture.
	\cite{xu2021rpvnet} utilizes the above three representations and proposes a range-point-voxel fusion network.
	Recently, \cite{yan20222dpass} applies cross-modal knowledge distillation, introducing prior information from texture and color images during the training phrase.
	Nevertheless, the hybrid-representation architecture makes them less robust in out-of-domain corruptions
	
	\begin{table*}
		\centering
		\caption{Categories and descriptions of corruptions in {our robustness benchmark.} We categorize common LiDAR corruptions into three domains: (1) adverse weather conditions, (2) measurement noise and (3) cross-device discrepancy. }
		\begin{tabular}{l|c|c}
			\toprule
			Corruption (C)           & Intensity (I) & Description                                                  \\\hline
			\multirow{3}{*}{(1) Fog Simulation}            & Light       & Fog simulation with $\beta=0.005$                                \\
			& Moderate  & Fog simulation with beta $\beta=0.06$                                \\
			& Heavy      & Fog simulation with beta $\beta=0.2$                                 \\\hline
			\multirow{3}{*}{(1)  Snowfall Simulation}       & Light       & Snowfall simulation with snowfall rate of 0.5mm/h                         \\
			& Moderate  & Snowfall simulation with snowfall rate of 1.5mm/h                       \\
			& Heavy      & Snowfall simulation with snowfall rate of 2.5mm/h                        \\\hline
			\multirow{3}{*}{(2)  Global Outliers} & Light       & 0.1\% extra noisy points uniformly in the 3D space                        \\
			& Moderate  &   5\% extra noisy points uniformly in the 3D space                             \\
			& Heavy      & 50\% extra noisy points uniformly in the 3D space                            \\\hline
			\multirow{3}{*}{(2)  Local Distortion}   & Light       & 20\% points with randomly jitter distortion $\sigma=0.05$   \\
			& Moderate  & 20\% points with randomly jitter distortion $\sigma=0.1$   \\
			& Heavy      & 20\% points with randomly jitter distortion $\sigma=0.2$     \\\hline
			\multirow{2}{*}{(3)  Cross 32-beam Device}        & Dense     & Reduce LiDAR beams to 32                                     \\
			& Sparse    & Reduce LiDAR beams to 32, sample 1/2 points in each beam  \\\hline
			\multirow{2}{*}{(3)  Cross 16-beam Device}        & Dense     & Reduce LiDAR beams to 16                                     \\
			& Sparse    & Reduce LiDAR beams to 16, sample 1/2 points in each beam \\\bottomrule
		\end{tabular}
		\label{corruption_class}
	\end{table*}
	
	\subsection{Robustness Benchmarks for Images}
	There are comprehensive robustness benchmarks for 2D image processing, spanning different tasks such as classification, semantic segmentation and instance segmentation.
	For robust image classification, ImageNet-C~\cite{hendrycks2019benchmarking} is the pioneer for these field, which corrupts the ImageNet~\cite{deng2009imagenet}’s test set with simulated corruptions such as motion blur, adverse weather and noises.
	After that, ObjectNet~\cite{barbu2019objectnet} build a benchmark with diverse corruptions in rotation, background and viewpoint, and ImageNetV2~\cite{recht2019imagenet} follow ImageNet and re-collects a test set to benchmark the robustness against natural distribution shift. 
	Recently, ImageNet-A and ImageNet-R are proposed by \cite{hendrycks2021natural}, which benchmarks classifier’s robustness against natural adversarial examples.
	Since ImageNet is initially proposed for diverse tasks, there also exists preliminary attempts to benchmark the robustness of model trained on ImageNet to other downstream tasks, such as semantic segmentation~\cite{kamann2020benchmarking}, instance segmentation~\cite{altindis2021benchmarking} and object detection~\cite{yamada2022does}.
	In the field of autonomous driving, there are also existing works producing corruptions on Cityscapes~\cite{Cordts2016Cityscapes}, \eg, investigating models' robustness against adverse weathers~\cite{porav2020rainy,sakaridis2018semantic} or other corruptions~\cite{michaelis2019benchmarking}.
	Recently, ACDC~\cite{sakaridis2021acdc} dataset collects four common adverse conditions in self-driving, \ie, fog, nighttime, rain, and snow, evaluating the models' robustness against these real-world corruptions.
	However, since the difference between 2D-3D data and model architecture, there is still huge demands of a comprehensive 3D robustness benchmark for semantic segmentation.

	\subsection{3D Robustness Benchmarks}
	In the field of autonomous driving, there lacks a robustness benchmark for \textbf{LiDAR semantic segmentation} to the best of our knowledge. 
	Existing surveys mostly focus on the point cloud classification task.
	For instance, \cite{xiao2021triangle} and \cite{zhang2022riconv++} propose disturbance and rotation invariant feature extraction, however, they cannot achieve state-of-the-art performance on the clean dataset.
	Other works boost models' robustness against adversarial corruptions by denoising and upsampling~\cite{zhou2019dup}, voting on subsampled point clouds~\cite{liu2021pointguard}, and applying local relative position~\cite{dong2020self}.
	There are robustness benchmarks for point cloud classification. 
	Specifically, RobustPointSet~\cite{taghanaki2020robustpointset} and PointCloud-C~\cite{ren2022modelnet-c} evaluate the robustness of point cloud classifiers under different corruptions.
	However, these approaches test robustness against corruptions purely on synthesis dataset, \ie, ModelNet40, and thus the obtained experience and conclusions are often unreliable in real-world self-driving applications.
	
	There are also investigations to improve the robustness on LiDAR scenarios. PointASNL~\cite{yan2020pointasnl} proposes adaptive sampling, which adaptive shifts the outlier points onto objects' surfaces, and thus boosts the robustness against noisy point clouds.
	%
	%
	In recent year, there are studies investigating performance of object detector in different adverse weathers, where they aim at mitigating the rarity of adverse weather effects.
	Specifically, \cite{hahner2021fog} and \cite{hahner2022lidar} independently propose fog and snowfall simulation, greatly boosting the robustness of object detection models on real-world adverse weathers.
	However, there is no real-world or simulated adverse weather data set for LiDAR semantic segmentation at present.
	Moreover, there exists preliminary attempts to investigate the robustness issue of the fusion methods for 3D object detection~\cite{bai2022transfusion,li2022deepfusion,yu2022benchmarking}. 
	Concretely, TransFusion~\cite{bai2022transfusion} evaluates the robustness of different fusion strategies under several scenarios, \eg, daytime and nighttime, DeepFusion~\cite{li2022deepfusion} test the model robustness by adding noise to LiDAR reflections and camera pixels and \cite{yu2022benchmarking} proposes a robust benchmark for LiDAR-camera fusion, which analyzes seven cases of robustness scenarios.  
	
	By contrast, we rigorously investigate the LiDAR system and identify three categories, in a total of 16 LiDAR corruptions for semantic segmentation, and develop a toolkit that transforms the existing dataset into a robustness benchmark. 
	We hope our study can boost future research to benchmark the robustness, and give researchers more insights about designing a robust semantic segmentation model.

	\section{Corruptions Taxonomy}\label{sec:benchmark}
	Real-world LiDAR scans can suffer from a wide range of corruptions, based on which we provide a taxonomy of the corruptions. 
	In this paper, we categorize common LiDAR corruptions into three domains, \ie, adverse weather conditions, measurement noise and cross-device discrepancy, in which we produce total six corruptions with 16 severity levels.
	By applying these six types of corruptions to SemanticKITTI~\cite{behley2019semantickitti}, {we generate two corrupted dataset, \ie, SemanticKITTI-C and SemanticPOSS-C, which is summarized in Tab.~\ref{corruption_class}.}
	In the remaining sections, we will introduce each corruption.
	
	\begin{figure}
		\centering
		\includegraphics[width=\linewidth]{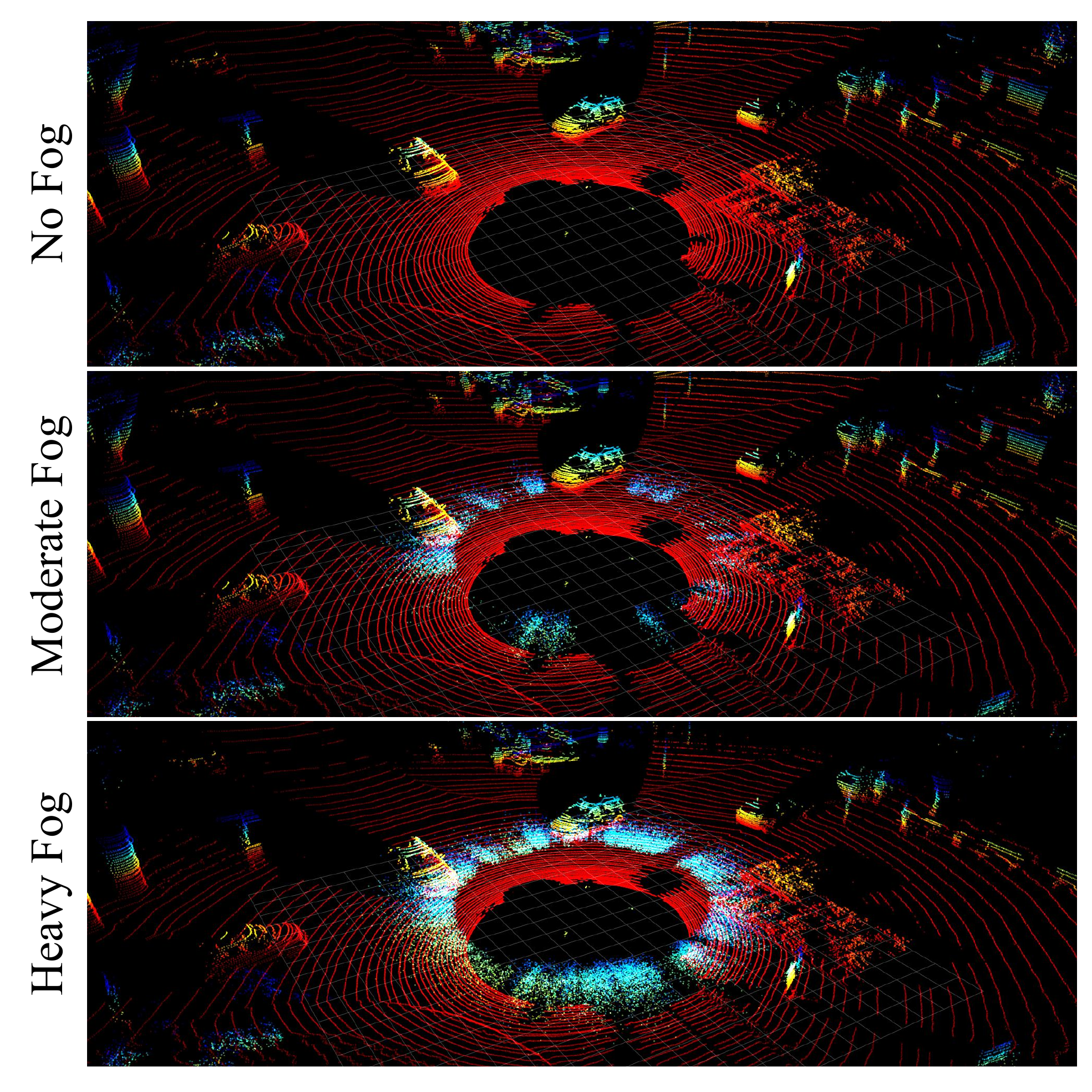}
		\caption{\textbf{Corruption of fog simulation.} We demonstrate the raw LiDAR point cloud in the first row. The foggy point clouds with  $\beta=0.06$ and $\beta=0.2$ are shown in the last two rows. The point cloud is color coded by the height (z value). The best viewed on a screen and zoomed in.}
		\label{fig:fog}
	\end{figure}
	
	A point cloud $\mathcal{P}$ is a set of points $\{p_j\}_{j=1}^N$, where $N$ is the number of points and $p_j \in \mathbb{R}^3$ includes the XYZ coordinates of the point $j$.
	A corruption operation is defined as a set-to-set function:
	\begin{equation}
		\begin{aligned}
			\mathcal{F}: \mathbb{R}^{N \times (3+D)} \mapsto \mathbb{R}^{N' \times (3+D)},
		\end{aligned}
	\end{equation}
	which maps the clean point cloud $\mathcal{P} = \{p_j\}_{j=1}^N$ and its $D$-dimensional features (if exist) to  corrupted ones (\eg, $\mathcal{P'} = \{p_j\}_{j=1}^{N'}$). 
	%
	%
	For LiDAR point cloud, each $p_j$ is associated with an intensity value $i_j \in \mathbb{R}$, indicating the return strength of a laser beam.
	%
	In this paper, the intensity is utilized to generate corrupted data, but we do not purely investigate the corruption of intensity.

	\begin{figure}
		\centering
		\includegraphics[width=\linewidth]{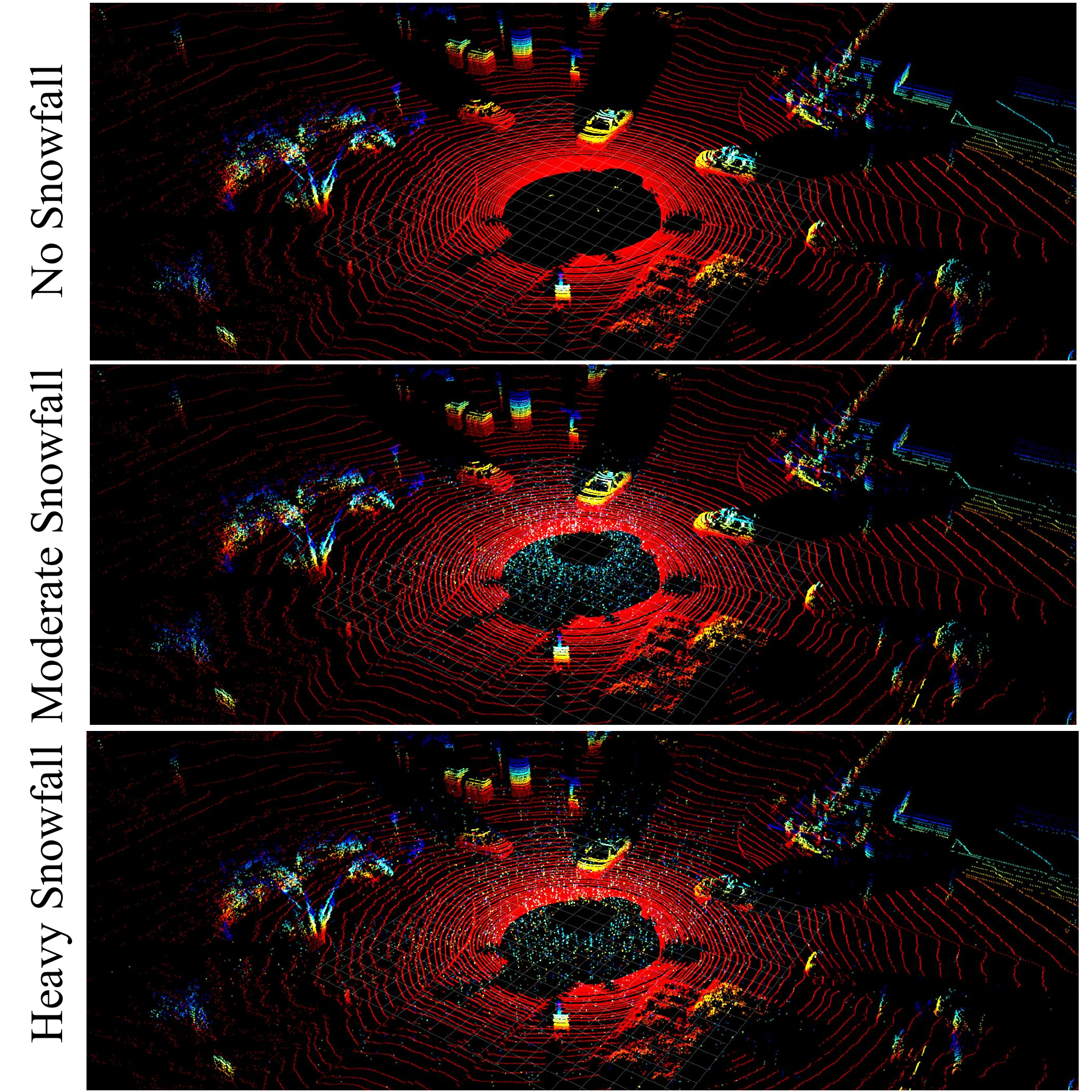}
		\caption{\textbf{Corruption of snowfall simulation.} We demonstrate the raw LiDAR point cloud in the first row. The snowfall point clouds with  snowfall rates 1mm/h and 2.5mm/h are illustrated in the last two rows. The point cloud is color coded by the height (z value). The best viewed on a screen and zoomed in.}
		\label{fig:snow}
	\end{figure}

	\subsection{Adverse Weather}
	In this section, we analyze two common weather conditions, namely fog and snowfall.
	%
	%
	For fog simulation, we follow ~\cite{hahner2021fog} to add fog to clean-weather point clouds by disturbing points' positions and intensities  according to physically valid rules. 
	Specifically, for a point ${p}\in \mathbb{R}^3$ captured in the clean weather, we first calculate its attenuated response $i_{\mathrm{hard}}$ in fog:
	\begin{equation}
		i_{\mathrm{hard}} = i \times \exp \left(-2 \alpha \times \lVert (x,y,z) \rVert\right),
	\end{equation}
	where $(x,y,z)$ is ${p}$'s coordinate in the ego frame and $i$ is its measured intensity, $\alpha$ is the attenuation coefficient in foggy weather, 
	$\lVert (x,y,z) \rVert$ denotes the distance between the point ${p}$ and the origin.
	Following ~\cite{hahner2021fog}, we uniformly sample $\alpha$ from [0, 0.005, 0.01, 0.02, 0.03, 0.06] when applying fog simulation to each sample.
	After that, we use the simulation terms in ~\cite{hahner2021fog} to compute the maximum fog response $i_{\mathrm{soft}}$ and its location $(x_{s},y_{s},z_{s})$, which lies in the line connecting the sensor and ${p}$. Note that the magnitude of $i_{\mathrm{soft}}$ is controlled by a backscattering coefficient $\beta$, which is manually set during the simulation.
	As shown in Tab.~\ref{corruption_class}, we choose $\beta$ from [0.005, 0.06, 0.2] to conduct fog simulation with different levels.
	Finally, the updated point position and its intensity are given by:
	\begin{equation}
		i =\left\{\begin{aligned}
			&i_{\mathrm{soft}} ~\text{if $i_{\mathrm{soft}} > i_{\mathrm{hard}}$},\\
			&i_{\mathrm{hard}} ~~ \text { otherwise. }
		\end{aligned}\right.
	\end{equation}
	\begin{equation}
		(x,y,z) =\left\{\begin{aligned}
			&(x_{s},y_{s},z_{s}) ~\text{if $i_{\mathrm{soft}} > i_{\mathrm{hard}}$},\\
			&(x,y,z) ~~ \text { otherwise. }
		\end{aligned}\right.
	\end{equation}
	In other words, if the fog is thick enough to overshadow the solid object point ${p}$ ($i_{\mathrm{soft}} > i_{\mathrm{hard}}$), we use the fog response (the intensity and position) to replace the original one. Otherwise, we keep the position of the original response with an attenuated intensity.
	
	The overall idea of this snowfall simulation is similar to that of the fog simulation.
	But unlike fog that homogeneously spreads in the 3D space, snowflakes are treated as opaque particles and are discretely distributed in space without intersecting with each other.
	For snowfall simulation, we follow ~\cite{hahner2022lidar} to sample snow particles for each LiDAR line and use them to modify the return for each LiDAR beam accordingly.
	The sampling function samples snow particles according to a given snowfall rate (mm/h), which controls the number of particles in a certain range.
	As shown in Tab.~\ref{corruption_class}, we separately set the snowfall rate to 0.5/1.5/2.5 to simulate light/moderate/heavy snowfall.

	The above two weathers have different characteristics.
	For instance, there are large areas of noisy points distributed around the sensor in foggy weather, and makes objects sparser due to the occlusion, especially for the remote objects.
	As illustrated in Fig.~\ref{fig:fog}, the number of these noisy points grows as the intensity of fog increases.
	Moreover, the noise introduced by fog is not uniformly distributed around the sensor. 
	The presence of noise depends on whether there is any object in the line of sight below a certain range from the sensor. 
	Generally, there will be few spurious returns from the respective pulses if a solid object exists at a moderate range.
	Inversely, if there is no object in a certain range, there are a lot of spurious returns that are caused by fog. 
	As for the snowfall, there are two explicit characteristics.
	On the one hand, the snow particles are explicitly modeled as opaque spheres, whose sizes are controlled by the snowfall rate.
	As shown in Fig.~\ref{fig:snow}, compared with foggy LiDAR where the noisy points are almost around the sensor, the noisy points in snowfall conditions are distributed more uniformly.
	Also, the snowfall rate does not greatly affect the number of noisy points but the size of snowy particles.
	On the other hand, wetness on the ground will exist in snowfall, where the emerging thin water layer increases the specular component of reflection by the ground surface. 
	To sum up, these two corruptions impact the models through global noisy points, making remote points sparser. 
	Nevertheless, they generally have different patterns.

	\subsection{Measurement Noise}\label{sec:noise}
	
\begin{figure}
		\centering
		\includegraphics[width=\linewidth]{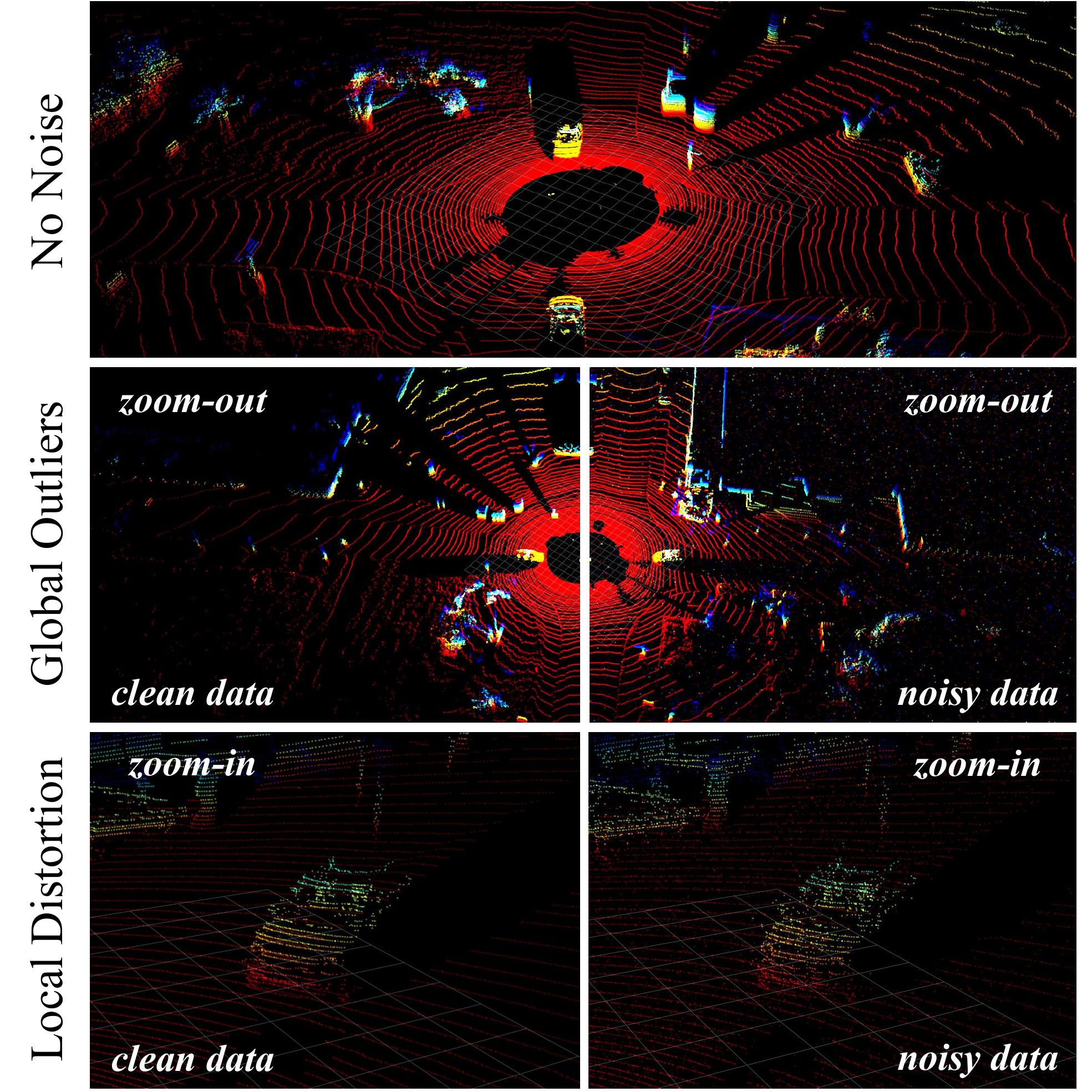}
		\caption{\textbf{Noisy LiDAR point clouds.} We demonstrate the raw LiDAR point cloud in the first row. The noisy point clouds with global outliers and local distortion are shown in the last two rows. The point cloud is color coded by the height (z value). The best viewed on a screen and zoomed in.}
		\label{fig:noise}
	\end{figure}

	Besides adverse weather, noises may also appear when corruption occurs during the data transmission or the sensors fail to capture information properly. 
	On one hand, due to the sensor overheated, false activations at max-distance makes a large number of noise points in the 3D space.  On the other hand, external disturbances, \textit{e.g.,} bumpy surfaces, dust, insects, that often lead to nonnegligible motion blur in local regions.
	We model such data disturbance using two types of random noises, as shown in Fig.~\ref{fig:noise}.
	Note that we only consider point coordinates during such corrupting operations.
	
	\noindent\textbf{Global outliers.} We randomly sample noises in a unit sphere and then merge them into a clean point cloud with proper rescaling.
	Such noises span the whole scene globally and are not conditional on the geometry of the clean point cloud. 
	Formally, given the clean point cloud $\mathcal{P} \in \mathbb{R}^{N \times 3} $, the corrupted point cloud $\mathcal{P}' \in \mathbb{R}^{N' \times 3} $ is obtained by:
	\begin{equation}
		\begin{aligned}
			\mathcal{P}' = \mathcal{P} \union \mathcal{P}_{noise},
		\end{aligned}
	\end{equation}
	where $\mathcal{P}_{noise} \in \mathbb{R}^{N^g\times 3}$ denotes sampled noises and $N' = N^g + N$. We control the noise intensity by selecting the proportion of noises $\frac{N^g}{N}$ from [0.1\%, 5\%, 50\%] as shown in Tab.~\ref{corruption_class}. 
	
	\noindent\textbf{Local distortion.} We randomly select some points within a scene and add Gaussian noises to their coordinates.
	Unlike global noises, which add additional points to a scene, local noises do not change the number of points.
	Compared to global noises, local noises jitter around a local neighborhood of the original points, mimicking the noisy disturbance during the data collection. 
	Formally, the local distortion point cloud $\mathcal{P}' \in \mathbb{R}^{N\times 3}$ is given by:
	\begin{equation}
		\begin{aligned}
			\mathcal{P}_{sub} = \text{RandomSample}(\mathcal{P},N^l),\\
			\mathcal{P}' = (\mathcal{P}_{sub} + \mathcal{O}) \union (\mathcal{P} \setminus \mathcal{P}_{sub}),
		\end{aligned}
	\end{equation}
	where $\text{RandomSample}(\cdot)$ randomly sample $N^l$ points from the clean point cloud $\mathcal{P}$. $\mathcal{O} \in \mathbb{R}^{N^l \times 3}$ denotes the random offsets sampled from a Gaussian distribution $\mathcal{N}(0, \sigma^2)$. $+$ and $\setminus$ are element-wise addition and set exclusion, respectively. 
	As shown in Tab.~\ref{corruption_class}, we choose $\sigma$ from [0.05, 0.1, 0.2] to control the jittering range in three different levels.
	{The proportion of noises is 20\%}

	\begin{figure}
		\centering
		\includegraphics[width=\linewidth]{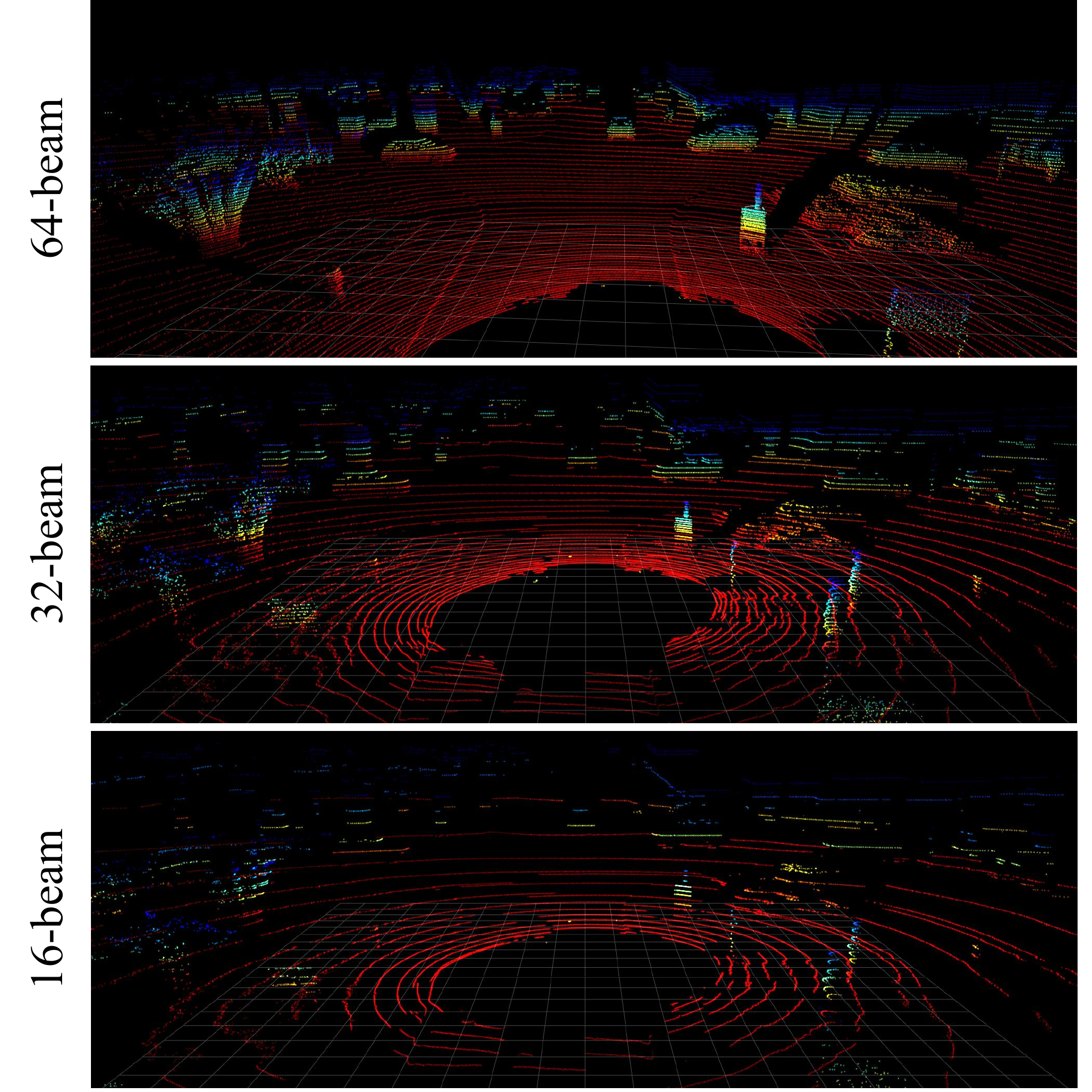}
		\caption{\textbf{Cross-device LiDAR point clouds.} We demonstrate the 64-beam LiDAR point cloud in the first row. The second  and third rows illustrate the 32-beam and 16-beam LiDAR data. The point cloud is color coded by the height (z value). The best viewed on a screen and zoomed in.}
		\label{fig:crossdevice}
	\end{figure}

	\begin{table*}[t]
		\centering
		\footnotesize
		\caption{LiDAR semantic segmentation approaches on our benchmark.}
		\resizebox{\textwidth}{!}{\begin{tabular}{cl|ccc}
				\toprule
				\multicolumn{1}{c}{Mainstream} & Method & Main representation & Extra representation & Reference \\\hline
				\multirow{4}[0]{*}{{Projection-based}} & SalsaNext~\cite{cortinhal2020salsanext} & Range image & -     & ArXiv 2020 \\
				& PolarNet~\cite{zhang2020polarnet} & BEV image& -     & CVPR 2020 \\
				& CENet~\cite{cheng2022cenet} & Range image & -     & ICME 2022 \\
				& GFNet~\cite{qiu2022gfnet} & Range and BEV images& Point cloud & TMLR 2022 \\\hline
				\multirow{3}[0]{*}{{Point-based}} & KPConv~\cite{Thomas_2019_ICCV} & Point cloud & -     & ICCV 2019 \\
				& RandLANet~\cite{hu2019randla} & Point cloud & -     & CVPR 2021 \\
				& Point Transformer~\cite{zhao2021point} & Point cloud & -     & ICCV 2021 \\\hline
				\multirow{4}[0]{*}{{Voxel-based}} & MinkowskiNet~\cite{choy20194d} & Grid voxel & -     & CVPR 2019 \\
				& SPVCNN~\cite{tang2020searching} & Grid voxel & Point cloud     & ECCV 2020 \\
				& Cylinder3D~\cite{zhou2020cylinder3d} & Cylinder voxel & Point cloud & CVPR 2021 \\
				& 2DPASS~\cite{yan20222dpass} & Grid voxel & Point cloud & ECCV 2022 \\\bottomrule	
		\end{tabular}}
		\label{tab:method}%
	\end{table*}%
	
	\subsection{Cross-Device Discrepancy}\label{sec:cross_device}
	An ideal segmentation algorithm is supposed to be robust across different devices with various specifications.
	While multiple factors (\eg, beam number and scanning speed of the LiDAR sensors) cause cross-device domain shifts, we focus on the beam number in this paper.
	To ensure high-quality data annotation, most large-scale datasets~\cite{geiger2012cvpr,sun2020scalability} are collected using high-resolution LiDARs.
	However, due to prohibitive costs, most practical vehicles are only shipped with low-beam sensors.
	For instance, KITTI~\cite{geiger2012cvpr} collects data through 64-beam LiDAR and each beam contains 1863 points in average, while those parameters in NuScenes~\cite{caesar2020nuscenes} are 32-beam and 1084 points.
	This suggests that an ideal segmentation model should be able to robust to different data distributions generated by different sensors. 
	%
	Unlike other factors introduced in previous subsections, the beam-induced domain gap is directly caused by the cross-device discrepancy instead of the collecting environment, making it also very important in our robustness analysis.

	To include the beam-based cross-device discrepancy in our benchmark dataset, we downsample the high-beam data (\ie, 64-beam) to low-beam data (\eg., 16-, 32-beam) using beam-level downsampling as shown in Fig.~\ref{fig:crossdevice}.
	One necessary information needed for beam-level downsampling is the beam label for each point, which is usually unknown for most datasets.
	To this end, we first assign a beam label to each point according to its zenith value in the spherical coordinate, which can be obtained via the following conversion:
	\begin{equation}\label{eqn:cart2sp}
		\begin{aligned}
			\theta=\arctan \frac{z}{\sqrt{x^2+y^2}},~~
			\phi=\arcsin \frac{y}{\sqrt{x^2+y^2}},
		\end{aligned}
	\end{equation}
	where $(x,y,z)$ is the Cartesian coordinate of the point and the $\theta$ and $\phi$ are zenith and azimuth angles.
	Following ~\cite{wei2022lidar}, we obtain beam labels by applying K-Means clustering on the zenith angles, where the number of clusters is set as the actual beam number of the high-beam point cloud.
	Compared to assigning beam labels by putting zenith angles into evenly distributed bins, the clustering-based technique does not require a pre-define zenith range and thus is more robust across different datasets.
	For a high-beam point cloud with the beam labels, we can easily downsample it into data with any lower beam number.
	In practice, we downsample point clouds with the beam numbers of 32 and 16.
	To simulate the diverse spinning speeds of the LiDAR devices, we evenly downsample points in each beam according to their azimuth angles. 
	By combining the above simulation, we have four corrupted data generated in Tab.~\ref{corruption_class}.

	\section{Candidate Methods}\label{sec:revisit}
	We benchmark 11 existing methods for LiDAR semantic segmentation, as shown in Tab.~\ref{tab:method}.
	Though we treat hybrid-representation methods as an independent mainstream in Sec.~\ref{sec:related_lidar}, current voxel-based and projection-based methods widely incorporate additional representations for auxiliary learning. Therefore, we categorize them only according to their main input representation, and the extra-representation will be illustrated if existed.

	\subsection{Projection-based Methods} 
	In this paper, we choose SalsaNext~\cite{cortinhal2020salsanext}, PolarNet~\cite{zhang2020polarnet}, CENet~\cite{cheng2022cenet} and GFNet~\cite{qiu2022gfnet} as the typical approaches of the projection-base method.
	These models project a LiDAR point cloud into 2D images and apply a 2D convolutional neural network for semantic segmentation. 
	Among the above methods, SalsaNext~\cite{cortinhal2020salsanext} and CENet~\cite{cheng2022cenet} conduct sphere projection to gain range views (RV), PolarNet~\cite{zhang2020polarnet} adopts polar projection to obtain bird’s-eye-view (BEV) under a polar coordinate system, and GFNet~\cite{qiu2022gfnet} uses the both.
	
\begin{figure}[t]
		\centering
		\includegraphics[width=\linewidth]{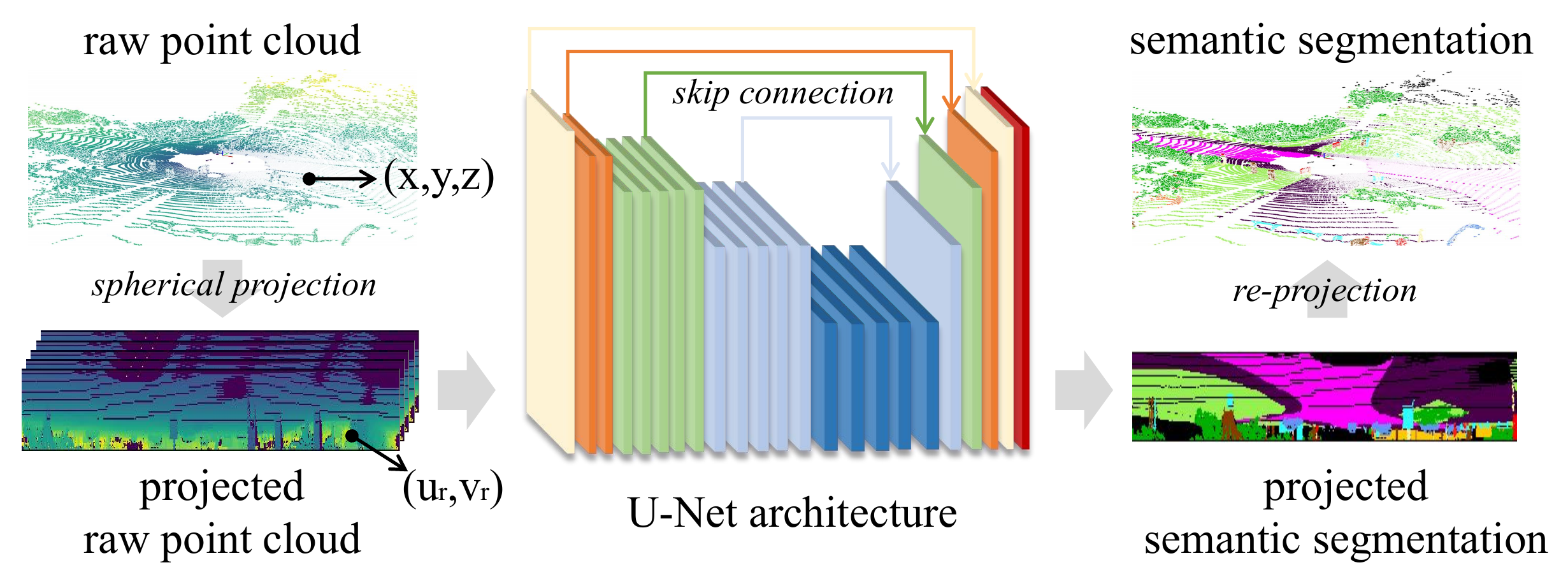}
		\caption{\textbf{Projection-based methods through spherical projection.} This kind of methods first conduct spherical projection maps each point to an image coordinate, and then adopt 2D convolution to construct a U-Net-like architecture. Finally, they re-project the prediction on the image plane onto the raw point cloud.}
		\label{fig:mainstream}
	\end{figure}

	\noindent\textbf{Sphere projection for range-view.} We denote $N$ the number of points in the LiDAR point cloud and $(H, W)$ are the height and width of the projected image. 
	As shown in Fig.~\ref{fig:mainstream}, the spherical projection maps each point to an image coordinate via
	\begin{equation}
		\binom{u_r}{v_r} = \binom{\frac{1}{2}[1 - \text{arctan}(y,x)\pi^{-1}]W}{[1 - \text{arsin}(zr^{-1}+fov_{up})fov^{-1}]H},
	\end{equation}
	where $p_i = (x, y, z)$ and $(u_r, v_r)$ are the $i$-th point and its coordinates on the range image plane. $r$ is the range of each point $\sqrt{x^2 + y^2 + z^2}$ and $fov = fov_{up} + fov_{down}$ is the vertical field-of-view of the sensor. 
	Finally, the LiDAR point cloud is converted to a \textbf{range image} with the shape of $(H \times W \times C)$.
	The channel $C$ is generally $5$, including $x$, $y$, $z$, $intensity$ and $range$ of the point.
	
	\noindent\textbf{Polar projection for bird’s-eye-view.}
	Existing methods also project the LiDAR point cloud to the bird’s-eye-view (BEV) through a top-down orthogonal projection.
	Considering the imbalanced spatial distribution in LiDAR data,  polar projection first transforms the BEV from the Cartesian system into a polar coordinate system through
	\begin{equation}
		\binom{u_p}{v_p} = \binom{\sqrt{x^2 + y^2 + z^2}\text{cos}(\text{arctan}(y, x))}{\sqrt{x^2 + y^2 + z^2}\text{sin}(\text{arctan}(y, x))},
		\label{polar}
	\end{equation}
	where $(u_p, v_p)$ is the coordinate transformation from the Cartesian system to the polar system.
	After that, they discretize $(u_p, v_p)$ to $[0, H - 1]$ and $[0, W - 1]$ and obtain a BEV image.

	\noindent\textbf{Architectures.}
	Since the LiDAR point cloud is already mapped onto an image plane, typical 2D semantic segmentation networks can be directly adopted.
	Specifically, U-Nets~\cite{ronneberger2015u} with specific modifications are applied in previous methods.
	1) Approaches with range images do not conduct pooling in height dimension due to the large width-height ratio of the input, as shown in Fig.~\ref{fig:mainstream}.
	2) SalsaNext utilizes an additional pixel-shuffle layer in the last encoder, and CENet conducts multiscale supervision in encoder and decoder layers.
	3) PolarNet applies a hybrid-representation manner, \ie, designing a PointHead (will be described in Sec.~\ref{sec:result3}) in the initial stage, which aggregates the features of original points into each BEV pixel, and finally conducts semantic segmentation through the 2D U-Net.
	4) GFNet~\cite{qiu2022gfnet} has a two-branch architecture, where two U-Nets independently encode the features of range-view (RV) and bird’s-eye-view (BEV). 
	There are several Geometric Flow (GF) modules between their decoder layers with different scales, which update each other's features by fusing the features of both branches. 
	Finally, it utilizes a hybrid-representation manner, aggregating the features of two branches in their last layers and feeding the fused feature into KPConv~\cite{Thomas_2019_ICCV} to gain point-wise predictions.
	
	\noindent\textbf{Configurations.} In our experiments, we adopt the official architectures of projection-based methods (\ie, SalsaNext\footnote{\url{https://github.com/TiagoCortinhal/SalsaNext}}, CENet\footnote{\url{https://github.com/huixiancheng/CENet}}, PolarNet\footnote{\url{https://github.com/edwardzhou130/PolarSeg}} and GFNet\footnote{\url{https://github.com/haibo-qiu/GFNet}}).
	Note that CENet is trained with a multi-stage strategy, which trains with $64\times 512$ range image for the initial stage, and fine-tunes the pre-trained model aggressively on $64\times 1024$ and $64\times 2048$ ones. Since the official codes {on SemanticKITTI} only provide the checkpoint on $64\times 512$ range images, we fine-tune the checkpoint on larger ones through provided configurations.
	In contrast, SalsaNext directly trains their model on $64\times 2048$ range images.
	As for the PolarNet, it first crops points of the polar coordinate system in the range from [3, $-\pi$, -3] to [50, $\pi$, 1.5], and then discretize points into a [480,360] BEV partition.
	GFNet combines RV and BEV representation, utilizing both  $64\times 2048$ range images and $480\times 360$ BEV plane.
	{The training setup on SemanticPOSS is the same as that on SemanticKITTI.}

	\begin{figure}
		\centering
		\includegraphics[width=\linewidth]{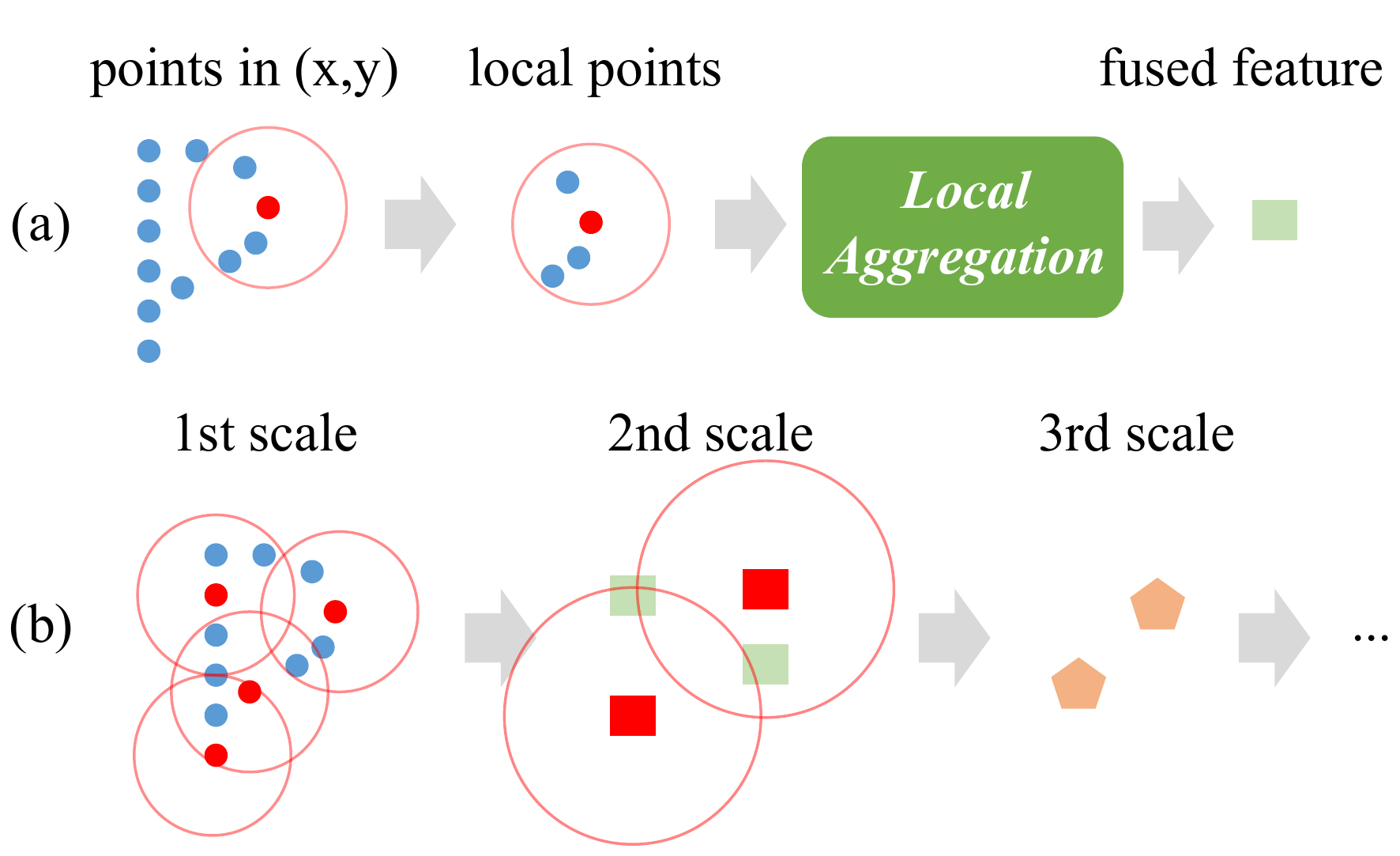}
		\caption{\textbf{Point-based methods.} (a) Point-based approaches sample the target points (in red color) in the original point cloud, and aggregate local features through local aggregation. (b) Through hierarchical architecture, the receptive field of the point-based method increase aggressively.}
		\label{fig:pointbase}
	\end{figure}

	\subsection{Point-based Methods} \label{pointbase}
	Point-based approaches aim at extracting features on raw point clouds directly, as shown in Fig.~\ref{fig:pointbase}.
	In this paper, KPConv~\cite{Thomas_2019_ICCV}, RandLA-Net~\cite{hu2019randla} and Point Transformer~\cite{zhao2021point} are selected as our candidate methods.
	Specifically, these methods first apply sampling approaches to select target points from the original point clouds, and then conduct local aggregation on each target point and mine local geometrics, as Fig.~\ref{fig:pointbase}(a) shows. 
	After constructing a hierarchical architecture in Fig.~\ref{fig:pointbase}(b), point-based methods gain the global semantic information of the input point cloud.
	
	\noindent\textbf{General formulation of local aggregation.} Let $p_i$ and $f_i$ denote the coordinate and feature of the $i$-th point.
	In general, for each $p_i$, a local aggregation function first transforms its neighbor $p_j$ with feature $f_j$ into a new feature by a transformation function $\mathcal{T}(...)$, and then aggregates all transformed neighborhood features to generate an updated feature of $\hat{p}_i$ via an aggregation function $\mathcal{A}(\cdot)$:
	\begin{equation}
		\hat{p}_i = \mathcal{A}(\{\mathcal{T}(p_i, f_i, p_j, f_j)\}~\forall j \in \mathcal{N}(p_i)).
		\label{eqn:localaggregation}
	\end{equation}
	In practice, $\mathcal{N}(p_i)$ represents the neighborhood index of point $p_i$. According to the category to which the transformation function $\mathcal{T}(...)$ belongs,
	previous local aggregation approaches can be roughly categorized into four classes: \textbf{1)}~{Point-wise MLP based}, \textbf{2)}~{Adaptive weight based},  \textbf{3)}~{Pseudo grid based} and \textbf{4)}~{Transformer based} approaches.
	The typical one of the first class is PointNet++~\cite{qi2017pointnet++}, where $\mathcal{T}$ and $\mathcal{A}$ are $\text{MLP}((p_j-p_i)\oplus   f_j)$ and max pooling respectively, in which $\oplus $ is concatenation operation. However, directly learn the 3D shapes through simple point-wise MLP and pooling cannot work well in the LiDAR scenario, \ie, it only achieves 20\% mIoU on SemanticKITTI in previous studies~\cite{behley2019semantickitti}. Therefore, we did not adopt this kind of methods in our paper.

	\noindent\textbf{Adaptive weight based methods.}
	The adaptive weight based methods design diverse convolution filters over arbitrary relative positions and hence compute weights on all neighbor points.
	RandLA-Net~\cite{hu2019randla} is a typical one in adaptive weight based methods.
	Concretely, its transformation function $\mathcal{T}$ can be represented as
	\begin{align}
		\text{MLP}(p_i \oplus p_j \oplus (p_i - p_j) \oplus \mathcal{E}(p_i, p_j))) \oplus f_j,
		\label{eqn:randla}
	\end{align}
	where $\mathcal{E}(\cdot)$ calculates the Euclidean distance between the neighboring and center points.
	After that, it aggregates the neighboring features through attention mechanism~\cite{vaswani2017attention}, which first calculates an attention weight according to the feature, and then conducts weighted average.
	
	\noindent\textbf{Pseudo grid based methods.} KPConv~\cite{Thomas_2019_ICCV} is a representative pseudo grid based method, which generates pseudo features on several sampled regular grid points, and thus regular convolution methods can play a normal role.
	Specifically, it samples equally distributed spherical grid points in the 3D space, in which the pseudo features $f^p_k$ on the $k$-th grid point can be calculated as
	\begin{align}
		f^p_k = \sum_{j \in \mathcal{N}(p_i)} \text{max}(0, 1-\frac{ \mathcal{E}(p_i, p_k)}{\sigma}) f_j,
	\end{align}
	where each grid point $p_k$ has a strict mapping with the relative position to the center point. $\sigma$ is a hyperparameter.
	After that, the transformation function $\mathcal{T}$ in pseudo grid based methods can be formulated as 
	\begin{align}
		\mathcal{T}(p_i, f_j) = w_k \odot  f^p_k,
	\end{align}
	where $w_k \in \mathbb{R}^{d\times 1}$ is a parametrized weight in the convolution operator and defined on each grid point.
	Finally, after applying max pooling as aggregation function $\mathcal{A}$, it updates the feature of each target point through aggregating features in local neighbors.

	\noindent\textbf{Transformer-based methods.}
	Besides analyzing traditional local aggregation based approaches, we also adopt a recent transformer-based method (\ie, Point Transformer~\cite{zhao2021point}) in this paper.
	The point transformer layer is based on vector self-attention, which uses the subtraction relation and there is a position encoding $\delta$ in both the attention vector $\gamma$ and the transformed features $\alpha$.
	Specifically, in each local group (\ie, $\forall j \in \mathcal{N}(p_i)$), the transformation function $\mathcal{T}(p_i, p_j, f_i, f_j)$ in a transformer based method can be formulated as 
	\begin{align}
		\rho(\gamma(\varphi(f_i)-\psi(f_j)+\delta)) \odot (\alpha(f_j) + \delta),
	\end{align}
	where $\varphi$, $\psi$, $\gamma$, $\alpha$ are independent MLPs. $\delta=\text{MLP}(p_i-p_j)$ is a positional encoding in self-attention, allowing the operator to adapt to local structure.
	After updating the neighboring features, Point Transformer utilizes a summation function as an aggregation function  $\mathcal{A}$ to fuse features.

	\noindent\textbf{Architectures.} 
	All above three methods follow the widely-used UNet-like encoder-decoder architecture with skip connections.
	The LiDAR point cloud is first fed to a shared MLP layer to extract per-point features. Encoder and decoder layers are then used to learn features for each point. Finally, fully-connected layers are used to predict the semantic label of each point.
	KPConv and RandLA-Net utilize stacked two corresponding local aggregation in each encoder layer, while Point Transformer using the combination of point-wise MLP with point transformer layer. In decoder layers, all method interpolate the sampled points and update features through point-wise MLPs.
	Moreover, RandLA-Net uses random sampling in each local aggregation, while other two utilizing uniformly sample points.
	For KPConv and RandLA-Net, we adopt their official architectures on SemanticKITTI dataset (four encoders and decoders). As for the Point Transformer, since it is only designed for indoor semantic segmentation, we adopt original architecture with five encoders and decoders.
	
	\noindent\textbf{Configurations.} 
	During the training, both KPConv\footnote{\url{https://github.com/HuguesTHOMAS/KPConv-PyTorch}} and RandLA-Net\footnote{\url{https://github.com/QingyongHu/RandLA-Net}} follow their official configurations {on SemanticKITTI}. 
	To accelerate the training phrase, they first conduct grid sampling with grid size 0.06$m$ to gain a small sub-cloud. Moreover, they respectively crop patches with a 4$m$ radius and 50,000 points in each training iteration. During the inference, they inference through small patches util each of the points have been inferred three times.
	Since there are not published codes on SemanticKITTI { and SemanticPOSS} for Point Transformer\footnote{\url{https://github.com/POSTECH-CVLab/point-transformer}}, we utilize the same configurations as \cite{tang2020searching} during the training and inference.

	\subsection{Voxel-based Methods} \label{voxelbase_revisit}
	Since voxel-based methods are the most popular mainstream for LiDAR semantic segmentation now, we select four methods (\ie, MinkowskiNet~\cite{choy20194d}, SPVCNN~\cite{tang2020searching}, Cylinder3D~\cite{zhou2020cylinder3d} and 2DPASS~\cite{yan20222dpass}) in this paper.
	
	\noindent\textbf{Grid partition.}
	Voxel-based methods exploit voxelization and transform the LiDAR point cloud into 3D voxels, such that the 3D convolutions can be applied.
	Specifically, they shift all the points to the local coordinate system with the geometric center as the origin. 
	Then, all the points are normalized into a unit sphere and scaled to the range of $[0, 1]$, where the normalized coordinates are denoted as $\hat{P} = \{(\hat{x}_i, \hat{y}_i, \hat{z}_i)\}_i^N$.
	After that, they transform the normalized point cloud to a voxel representation with voxel size $vs$ ($f_m^*$ is voxelized feature representation):
	\begin{equation}
		\begin{aligned}
			p_i^* &= ({x}_i^*,{y}_i^*,{z}_i^*) 
			= (\lfloor\hat{x}_i / vs\rfloor, \lfloor\hat{y}_i / vs\rfloor, \lfloor\hat{z}_i / vs\rfloor),\\
			f_m^* &= \frac{1}{N_m} \sum_{i=1}^N \mathbb{I}[{x}_i^* = \hat{x}_m, {y}_i^* = \hat{y}_m, {z}_i^* = \hat{z}_m] \cdot p_i,
			\label{vx}
		\end{aligned}
	\end{equation}
	where $\lfloor \cdot \rfloor$ is the floor function, and $\mathbb{I}(\cdot)$ is a binary indicator of whether ${p}_i^*$ belongs to the $m$-th voxel grid or not.
	$N_m$ is the number of points in the $m$-th voxel, and the original point coordinates are averaged as the features of each voxel.
	After the operations in Eqn.~\eqref{vx}, only the non-empty voxels are preserved ($N_m > 0$) in a hash table.
	The, the convolution operation only conducts on the non-empty voxels, thus maintaining the computational efficiency.
	
	\noindent\textbf{Cylindrical partition.}
	Recent study \cite{zhou2020cylinder3d} proposes cylinder partition for voxelization, which makes a higher non-empty proportion and more balanced point distribution compared with grid partition, especially for farther-away regions.
	In practice, it first transforms the Cartesian system into a polar coordinate system through Eqn.~\eqref{polar}, and then conducts voxelization as Eqn.~\eqref{vx}.

	\noindent\textbf{Architectures.} 
	Both MinkowskiNet and SPVCNN\footnote{\url{https://github.com/mit-han-lab/spvnas}} utilize the same U-Net architecture, where the difference is that there is a parallel point-wise MLP branch in the latter.
	Cylinder3D\footnote{\url{https://github.com/xinge008/Cylinder3D}} proposes asymmetrical 3D convolution networks, in which it constructs several asymmetrical blocks (\eg, exploiting $3\times 3\times 1$, $3\times 1\times 3$ and $1\times 3\times 3$ kernels in parallel) as unit components.
	2DPASS\footnote{\url{https://github.com/yanx27/2DPASS}} uses a similar encoder architecture as SPVCNN, but it discards the decoder part and predicts the results through multiscale concatenation.
	Moreover, Cylinder3D and 2DPASS exploit additional a PointHead (will be introduced in Sec.~\ref{sec:result3}) to aggregate point-wise features into each voxel.
	
	\noindent\textbf{Configurations.} 
	{Both Cylinder3D and 2DPASS are tested on SemanticKITTI-C with their released checkpoints. For the SemanticPOSS-C, and retrain them with the same setup as SemanticKITTI}.
	As for the MinkowskiNet and SPVCNN, we re-trained their official architectures with batch size 8 for epoch 64, and gain higher results.
	All approaches are tested with test-time augmentation (TTA), \ie, rotating the point cloud with 12 views and averaging the predictions.

	\section{Benchmarking and Analysis}\label{sec:methods}
	In this section, we benchmark the aforementioned  $11$ approaches with our diverse set of LiDAR corruptions. We first introduce the experiment setting and evaluation metrics of our benchmark in Sec.~\ref{sec:setting}.
	After that, the benchmark results are shown in Sec.~\ref{sec:result} to \ref{sec:result4} with comprehensive analysis.
	We demonstrate our benchmark results spanning different representation, architecture, corruption intensity and data augmentation.
	As results, we summarize \textbf{12 observations} in total.
	Finally, in Sec.~\ref{sec:ours}, we introduce RLSeg, a robust architecture based on the above observations, which effectively boosts the robustness of LiDAR semantic segmentation. 
	
	\subsection{Experiment Setting}\label{sec:setting}
	\noindent\textbf{Dataset.} SemanticKITTI is currently the most widely used LiDAR semantic segmentation dataset, which consists of 43,552 densely annotated LiDAR scans belonging to 21 sequences. These scans are annotated with a total of 19 valid classes, and each scan spans up to $160\times160\times20$ meters with more than $\sim 10^5$ points.
	Initially, the sequence 00 to 07, 09 to 10 are the training set, 11 to 21 are the test set, and 08 is the validation set. Since the annotations of 11 to 21 are not available offline, we train all approaches on training set and evaluate them on sequence 08.
	
	{{SemanticPOSS} contains 11 similar annotated categories with SemanticKITTI. It is more challenging because each scene contains more than 10$\times$ sparse small objects (\textit{i.e.,} people and bicycle), while the total frames number are only 1/20 of SemanticKITTI. 
		We train all approaches on its training set (sequence 00, 01 and 03-05) and evaluate them on sequence 02.}
	
	\noindent\textbf{Annotation modification.} Since the corrupted point clouds will be sparser or there are new noisy points existed, we slightly modify the original annotations.
	Specifically, for the corruption data in fog and snowfall simulations, {we utilize the noisy data to query {the original point} of the clean data to annotate labels,  labeling points as `ignore' if there is no point existed.}
	In the noisy corruption (\ie, local and global), we directly annotate noisy points as `ignore'.
	In the cross-device scenario, since all the points are sampled from the original LiDAR, there is no demand for modification.
	\textbf{Note that the `ignore' class is not considered in the evaluation.}

	\noindent\textbf{Evaluation metric.}
	To intuitively demonstrate the robustness of candidate methods, we use the performance on each corruption and the relative performance degradation compared to the clean data on our benchmark datasets as our evaluation metrics.
	Specifically, we adopt mIoU(\%) (\ie, averaged Intersection over Union on each class) as our metric and the score on the clean dataset is denoted as $S$.
	As demonstrated in Tab.~\ref{corruption_class}, we benchmark existing methods with six categories of corruptions (C), spanning 16 different intensities (I).
	The performance toward certain corruption $c\in C$ can be calculated by averaging results on each intensity:
	\begin{align}
		S^c = \sum_{i \in I(c)} S^c_i / N(c),
	\end{align}
	where $I(c)$ and $N(c)$ are total intensities and the number of intensity in the corruption $c$. 
	$S_i^c$ denotes the mIoU under corruption $c$ and intensity $i$ and $S^c$ are the averaged mIoU of all intensities under the corruption $c$.
	The relative mean robustness performance of the model is defined as $\text{R}^c = S^c /S$. The higher R means the model is more robust to inferior LiDAR conditions.
	Moreover, we define a robustness mIoU (\textbf{RmIoU}) and averaged relative performance (\textbf{mR}) through averaging the results on different corruption:
	\begin{align}
		\text{RmIoU} = \sum_{c \in C} S^c / 6,  ~~ \text{mR} = \text{RmIoU} / S.
	\end{align}

	\begin{sidewaystable*}
		\centering
		{		
			\caption{Benchmarking the robustness of state-of-the-art methods in all 16 scenarios (6 classes) on SemanticKITTI-C. R denotes the relative mean robustness performance of the model. The higher R means the model is more robust to inferior LiDAR conditions.}
			\begin{tabular}{cl|c|cc|cccc|cccc|cccc}
				\toprule
				&       & \textit{Clean} & \multicolumn{2}{c|}{\textit{Robustness}} & \multicolumn{2}{c}{\textit{Fog}} & \multicolumn{2}{c|}{\textit{Snowfall}} & \multicolumn{2}{c}{\textit{Global Outliers}} & \multicolumn{2}{c|}{\textit{Local Distortion}} & \multicolumn{2}{c}{\textit{32-beam}} & \multicolumn{2}{c}{\textit{16-beam}} \\
				&   Method    & {mIoU} & {RmIoU} & {mR} & {mIoU} & {R} & {mIoU} & {R} & {mIoU} & {R} & {mIoU} & {R} & {mIoU} & {R} & {mIoU} & {R} \\\hline
				\multirow{4}[0]{*}{\rotatebox{90}{{Projection}}} & {SalsaNext} & 55.8 & 42.7 & 76.5 & 27.3 & 48.9 & 43.6 & 78.1 & 49.5 & 88.7 & 53.6 & 96.1 & 51.1 & 91.6 & 31.0 & 55.6 \\
				& {PolarNet} & 58.2 & 43.3 & 74.5 & 31.7 & 54.5 & 47.4 & 81.5 & 52.4 & 90.0 & 48.7 & 83.7 & 46.3 & 79.5 & 33.5 & 57.6 \\
				& {CENet} & 62.3 & 47.7 & 76.6 & 31.5 & 50.5 & 51.3 & 82.4 & 57.2 & 91.7 & 58.6 & 94.0 & 54.5 & 87.5 & 33.4 & 53.7 \\
				& {GFNet} & 63.0 & 46.4 & 73.6 & 31.1 & 49.4 & 41.8 & 66.3 & 61.4 & 97.5 &  56.0     &   88.9    &   52.5    &    83.4   &    35.6   & 56.5 \\\hline
				\multirow{3}[0]{*}{\rotatebox{90}{{Point}}} & {KPConv} & 63.5 & 51.6 & 81.3 & \textbf{59.6}  & 93.9 & 54.8 & 86.4 & 61.9 & 97.4 & 31.8 & 50.1 & 58.3 & 91.7 & 43.4 & 68.3 \\
				& {RandLA-Net} & 59.2 &47.6 & 80.4 & 56.4 & \textbf{95.4} &   50.0    &   84.4    & 57.8 & 97.8 & 26.4 & 44.7 &    53.3   &   90.1    &    41.7   & 70.5 \\
				& {Point Trans.} & 63.3 & 40.5 & 64.0 & 45.5 & 71.9 & 44.2 & 69.9 & 38.8 & 61.4 & 39.3 & 62.2 & 47.2 & 74.6 & 27.9 & 44.2 \\\hline
				\multirow{4}[0]{*}{\rotatebox{90}{{Voxel}}} & {MinkowskiNet} & 66.3 & 53.6 & 80.9 & 56.3 & 84.9 & 50.4 & 76.1 & 65.3 & 98.5 & 37.0 & 55.9 & 62.2 & \textbf{93.9} & \textbf{50.4} & \textbf{76.0} \\
				& {SPVCNN} & 67.4 & 53.2 & 78.9 & 53.7 & 79.7 & 50.5 & 75.0 & 65.8 & 97.7 & 39.6 & 58.8 & 61.7 & 91.5 & 47.7 & 70.8 \\
				& {Cylinder3D} & 66.9 & 46.5 & 69.5 & 44.2 & 66.1 & 45.7 & 68.3 & 63.3 & 94.6 & 39.7 & 59.4 & 51.2 & 76.5 & 34.7 & 51.9 \\
				& {2DPASS} & 70.1 & 51.1 & 72.9 & 40.4 & 57.6 & 53.6 & 76.5 & 69.8 & 99.6 & 43.9 & 62.7 & 61.3 & 87.4 & 37.7 & 53.7 \\\hline
				& RLSeg (ours) & \textbf{73.5} & \textbf{62.5} & \textbf{85.0} & {57.6}  & {78.4} &\textbf{66.2} & \textbf{90.1} & \textbf{73.4} & \textbf{99.9} &\textbf{71.9} &\textbf{97.8} & \textbf{62.3}&{84.7} &{43.6}& {59.3} \\ \bottomrule
			\end{tabular}
			\label{tab:benchmark}
			\vspace{.6cm}
			
			\caption{{{Benchmarking the robustness of state-of-the-art methods in all 16 scenarios (6 classes) on SemanticPOSS-C.} R denotes the relative mean robustness performance of the model. The higher R means the model is more robust to inferior LiDAR conditions. $\dagger$Since there is no corresponding images in SemanticPOSS, we did not apply cross-modal knowledge transfer in 2DPASS.}}
			\resizebox{\textwidth}{!}{\begin{tabular}{cl|c|cc|cccc|cccc|cccc}
					\toprule
					&& \textit{Clean} & \multicolumn{2}{c|}{\textit{Robustness}} & \multicolumn{2}{c}{\textit{Fog}} & \multicolumn{2}{c|}{\textit{Snowfall}} & \multicolumn{2}{c}{\textit{Global Outliers}} & \multicolumn{2}{c|}{\textit{Local Distortion}} & \multicolumn{2}{c}{\textit{32-beam}} & \multicolumn{2}{c}{\textit{16-beam}} \\
					&Method    & {mIoU} & {RmIoU} & {mR} & {mIoU} & {R} & {mIoU} & {R} & {mIoU} & {R} & {mIoU} & {R} & {mIoU} & {R} & {mIoU} & {R} \\\hline
					\multirow{4}[0]{*}{\rotatebox{90}{{Projection}}}&SalsaNext        & 49.2          & 31.1          & 63.1          & 21.9          & 44.6          & 6.4           & 13.0          & 37.7          & 76.6           & 48.0          & 97.6          & 42.0          & 85.3          & 30.4          & 61.8          \\
					&PolarNet         & 53.5          & 36.8          & 68.7          & 30.0          & 56.1          & 9.8           & 18.4          & 51.3          & 95.9           & 51.2          & 95.6          & 44.5          & 83.2          & 33.7          & 63.0          \\
					&GFNet            & 49.0          & 36.2          & 73.9          & 28.2          & 57.4          & 12.9          & 26.3          & 48.9          & 99.9           & 47.5          & 97.0          & 44.1          & 90.1          & {35.7} & \textbf{72.8} \\
					&CENet            & 50.9          & 34.5          & 67.8          & 35.5          & 69.8          & 4.5           & 8.8           & 42.1          & 82.8           & {50.2} & \textbf{98.6} & 44.0          & 86.4          & 30.6          & 60.1          \\\hline
					\multirow{3}[0]{*}{\rotatebox{90}{{Point}}} &KPConv           & 56.3          & 46.9          & {83.3} & \textbf{49.3} & 87.6          & 46.8          & 83.2          & {56.3} & \textbf{100.0} & 45.6          & 81.0          & 49.5          & 87.9          & 34.0          & 60.3          \\
					&RandLA-Net       & 51.2          & 42.5          & 83.0          & {46.4} & \textbf{90.6} & {42.8} & \textbf{83.6} & 49.4          & 96.5           & 42.2          & 82.4          & 42.5          & 82.9          & 31.8          & 62.1          \\
					&PointTransformer & 57.0          & 35.9          & 63.0          & 34.9          & 61.2          & 19.4          & 34.1          & 52.7          & 92.4           & 48.3          & 84.7          & 37.7          & 66.2          & 22.6          & 39.6          \\\hline
					\multirow{4}[0]{*}{\rotatebox{90}{{Voxel}}} & MinkowskiNet         & 57.9          & 44.5          & 76.8          & 39.8          & 68.8          & 23.4          & 40.3          & 55.0          & 94.9           & 55.4          & 95.6          & 53.0          & 91.5          & 40.3          & 69.5          \\
					&{SPVCNN}  & \textbf{60.4} & 45.7          & 75.7          & 41.7          & 68.9          & 20.3          & {33.5} & \textbf{60.4} & \textbf{100.0}          & 55.7          & {92.2} & \textbf{55.5} & \textbf{91.8} & \textbf{40.8} & 67.5          \\
					&Cylinder3D       & 48.7          & 31.1          & 63.9          & 28.4          & 58.5          & 11.8          & 24.3          & 46.4          & 95.3           & 44.3          & 91.0          & 32.8          & 67.4          & 22.9          & 47.1          \\
					&2DPASS$\dagger$           & 55.9          & 41.7          & 74.7          & 35.9          & 64.3          & 15.5          & 27.8          & 55.8          & 99.8           & 53.9          & 96.5          & 50.1          & 89.7          & 39.0          & 69.8          \\\hline
					&{RLSeg (ours)}            & {59.5} & \textbf{50.1} & \textbf{84.2} & 47.9          & {80.5} & \textbf{48.5} & 81.6          & 59.4          & {99.9}  & \textbf{58.5} & 98.4          & 51.3          & 86.2          & 34.9          & 58.6   \\\bottomrule     \label{tab:benchmark2} 
		\end{tabular}}}
	\end{sidewaystable*}%

	\begin{figure}
		\centering
		\includegraphics[width=\linewidth]{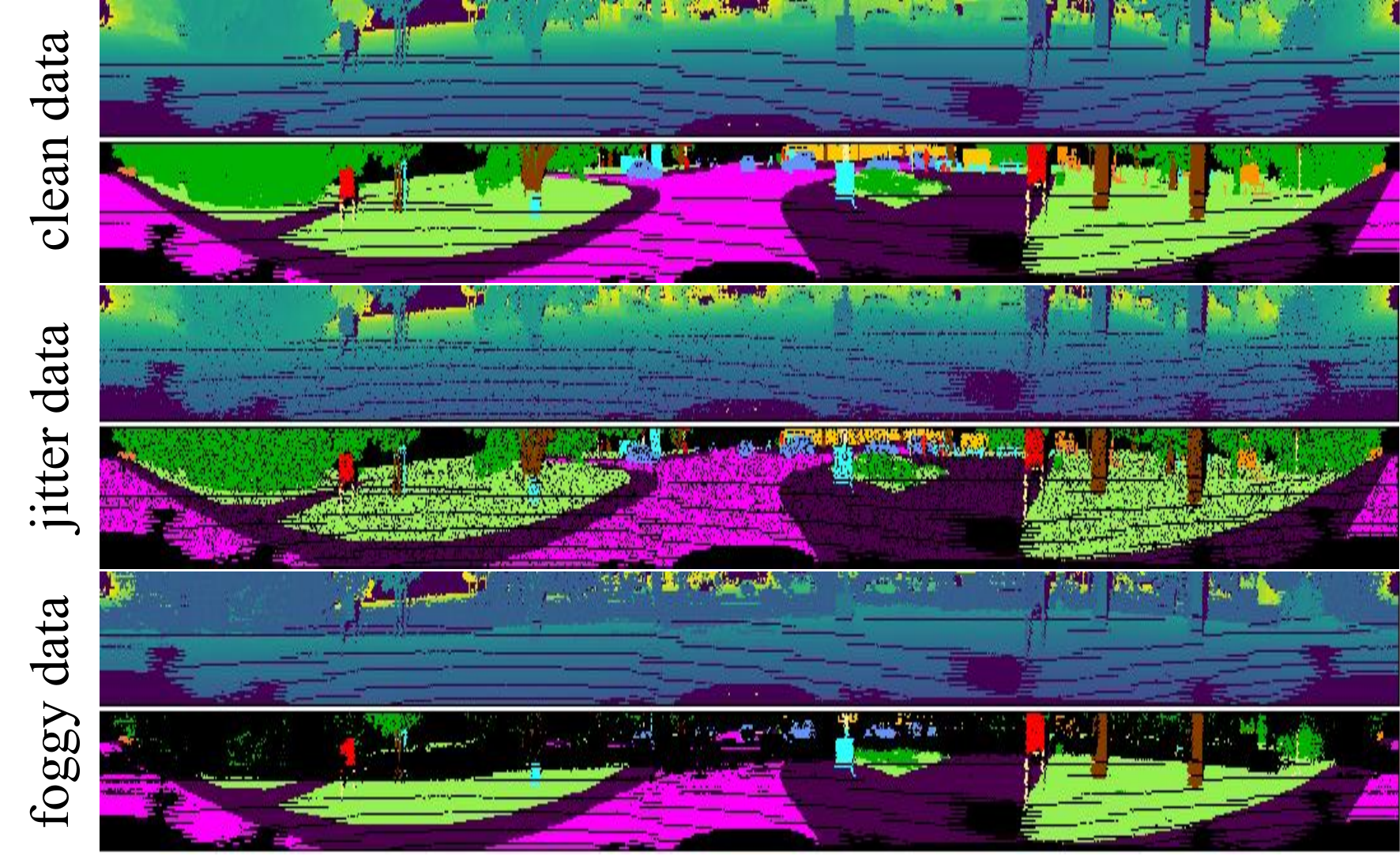}
		\caption{\textbf{Illustration of observation-1.} We show visualization results to better explain why the projection-based methods are vulnerable to fog simulation but robust to local deviation.}
		\label{fig:range_corruption}
	\end{figure}	
	
	\subsection{Main Results}\label{sec:result}
	Benchmark results are reported in Tab.~\ref{tab:benchmark} and \ref{tab:benchmark2}, in which projection-based, point-based and voxel-based methods are demonstrated in upper, median and lower parts, respectively. Our proposed solution will be introduced in Sec.~\ref{sec:ours}. According to the table, we have the following discovery:
	
	\noindent\textbf{Observation-1:} \textit{Projection-based methods are most vulnerable to common corruptions, especially to foggy simulation.  However, they are greatly robust to local distortion corruption.}
	
	As shown in the tables, {existing projection-based methods respectively achieve around 75\% and {70\%} metric of mR on SemanticKITTI-C and {SemanticPOSS-C,} which is much lower than those of point-based and voxel-based methods.
		Furthermore, they only achieve around 50\% original performance in fog simulation of SemanticKITTI-C and less than 20\% in the case of snow simulation on SemanticPOSS-C.}
	In contrast, these methods are extremely robust to local noise, especially for the pure range image based method (\ie, 96.1\% and 94.0\% R for SalsaNext~\cite{cortinhal2020salsanext} and CENet~\cite{cheng2022cenet} in SemanticKITTI-C, respectively). 
	The reason of the above observation is that the local corruption only slightly affects the range image and the foggy one make range images messy, as shown in Fig.~\ref{fig:range_corruption}.
	PolarNet~\cite{zhang2020polarnet} and GFNet~\cite{qiu2022gfnet} are not much robust to local corruption, since there are BEV projections in their models. Nevertheless, they still respectively keep 83.7\% and 88.9\% performance, which is much higher than those of point and voxel based methods.

	\begin{figure}[t]
		\centering
		\includegraphics[width=\linewidth]{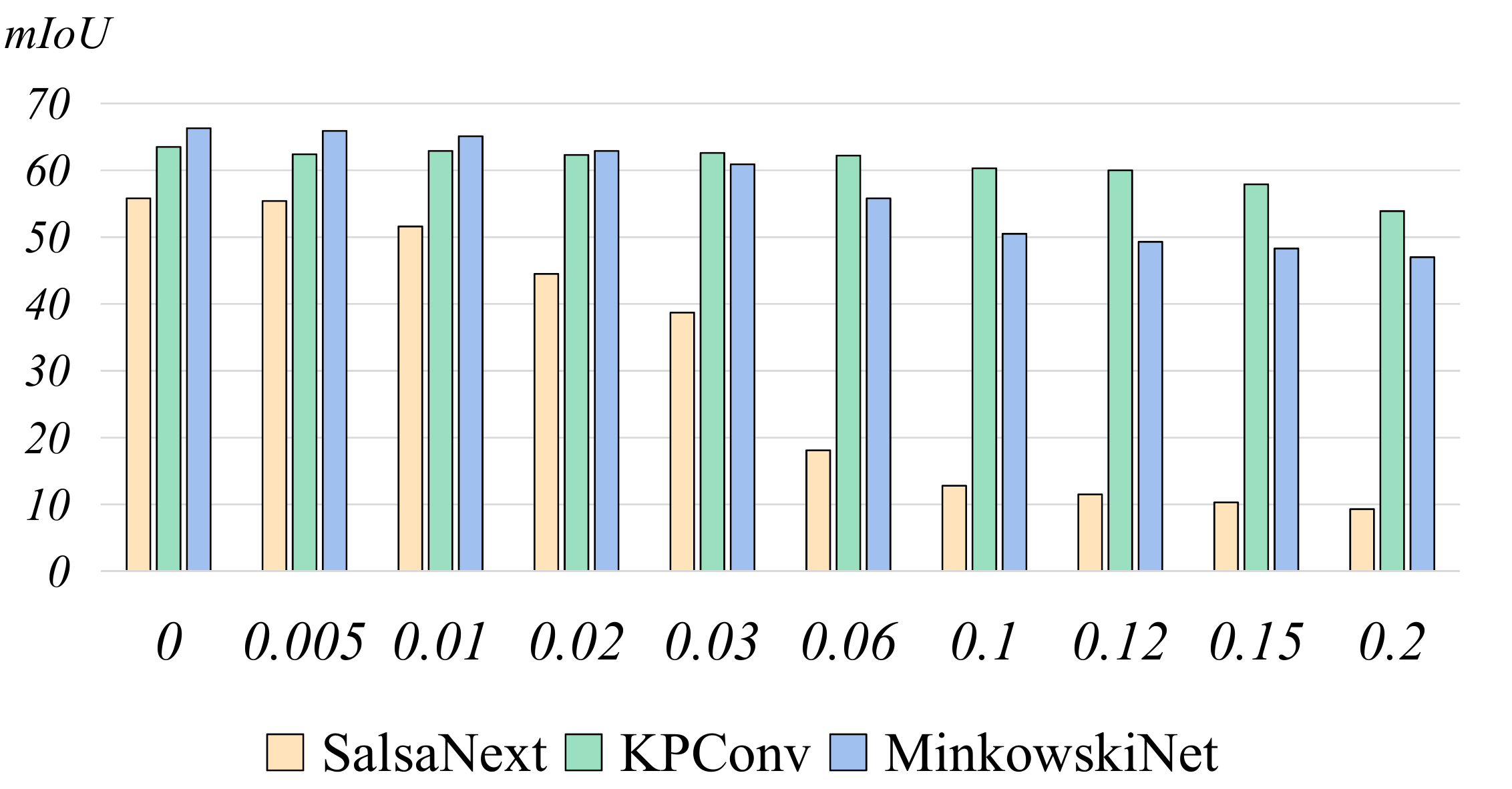}
		\caption{\textbf{Detailed results in fog simulation.} Performance of three typical approaches in different fog simulation intensities, where the x-axis denotes different $\beta$ values.}
		\label{fig:fog_results}
	\end{figure}
	
	\noindent\textbf{Observation-2:} \textit{Traditional point-based methods (RandLA-Net and KPConv) are more robust to common corruptions, compared with projection-based methods. Specifically, they are much robust to adverse weathers, but less robust to local distortion. }
	
	{The Tab.~\ref{tab:benchmark} and \ref{tab:benchmark2} illustrate RandLA-Net~\cite{hu2019randla} and KPConv~\cite{Thomas_2019_ICCV} respectively achieve {81.3\%/83.3\% and 80.4\%/83.0\%} in the metric of mR. Especially, they gain highest 93.9\% and 95.4\% R in the case of fog simulation on SemanticKITTI-C, where the best projection and voxel based methods only achieve 54.5\% and 79.7\%, respectively.}
	Inversely, they only gain 50.1\% and 44.7\% performance in local noise {of SemanticKITTI-C,} which is lower than common performance of other two mainstreams. 
	The reason is that the local distortion protects the local geometric and thus makes the local aggregation failed.
	
	\begin{table*}[t]
		\centering
		\caption{Comprehensive results on diverse noisy corruptions.}
		\begin{tabular}{lc|cc|cc|cc}
			\toprule
			\multirow{2}{*}{Noise types} & \multirow{2}{*}{Ratio (\%)} & \multicolumn{2}{c|}{SalsaNext} & \multicolumn{2}{c|}{KPConv} & \multicolumn{2}{c}{MinkowskiNet}  \\
			&                        & mIoU & R                      & mIoU & R                   & mIoU & R                          \\\hline
			No corruption                       & 0                      & 55.8 & 100.0                    & 63.5 & 100.0                 & \textbf{66.3} & 100.0                        \\\hline
			\multirow{5}{*}{Global outliers}               & 0.1                    & 55.8 & 100.0                    & 62.6 & 98.6                & \textbf{66.5} & 100.0                        \\
			& 5                      & 53.8 & 96.4                   & 62.8 & 98.9                & \textbf{65.9} & 99.4                       \\
			& 10                     & 51.6 & 92.5                   & 62.4 & 98.3                & \textbf{64.5} & 97.3                       \\
			& 20                     & 47.4 & 84.9                   & 61.8 & 97.3                & \textbf{63.9} & 96.4                       \\
			& 50                     & 38.9 & 69.7                   & 60.2 & 94.8                & \textbf{63.5} & 95.8                       \\\hline
			\multirow{3}{*}{Local distortion ($\sigma^2=0.05$)}           & 10                     & 55.7 & 99.8                   & 59.8 & 94.2                & \textbf{62.1} & 93.6                       \\
			& 20                     & 55.2 & 98.9                   & 35.5 & 55.9                & \textbf{55.5} & 83.7                       \\
			& 50                     & \textbf{50.7} & 90.9                   & 17.1 & 26.9                & 36.9 & 55.7                       \\\hline
			\multirow{3}{*}{Local distortion ($\sigma^2=0.1$)}          & 10                     & \textbf{55.4} & 99.3                   & 41.6 & 65.5                & 52.3 & 78.9                       \\
			& 20                     & \textbf{53.7} & 96.2                   & 25.6 & 40.3                & 32.3 & 48.7                       \\
			& 50                     & \textbf{42.7} & 76.5                   & 11.1 & 17.5                & 17.5 & 26.4                       \\\hline
			\multirow{3}{*}{Local distortion ($\sigma^2=0.2$)}           & 10                     & \textbf{54.8} & 98.2                   & 34.1 & 53.7                & 42.5 & 64.1                       \\
			& 20                     & \textbf{51.9} & 93.0                & 21.2 & 33.4                & 23.8 & 35.9                       \\
			& 50                     & \textbf{35.4} & 63.4                   & 14.1 & 22.2                & 11.2 & 16.9                  \\\bottomrule    
		\end{tabular}
		\label{tab:noisy}
	\end{table*}

	\begin{figure}[t]
		\centering
		\includegraphics[width=\linewidth]{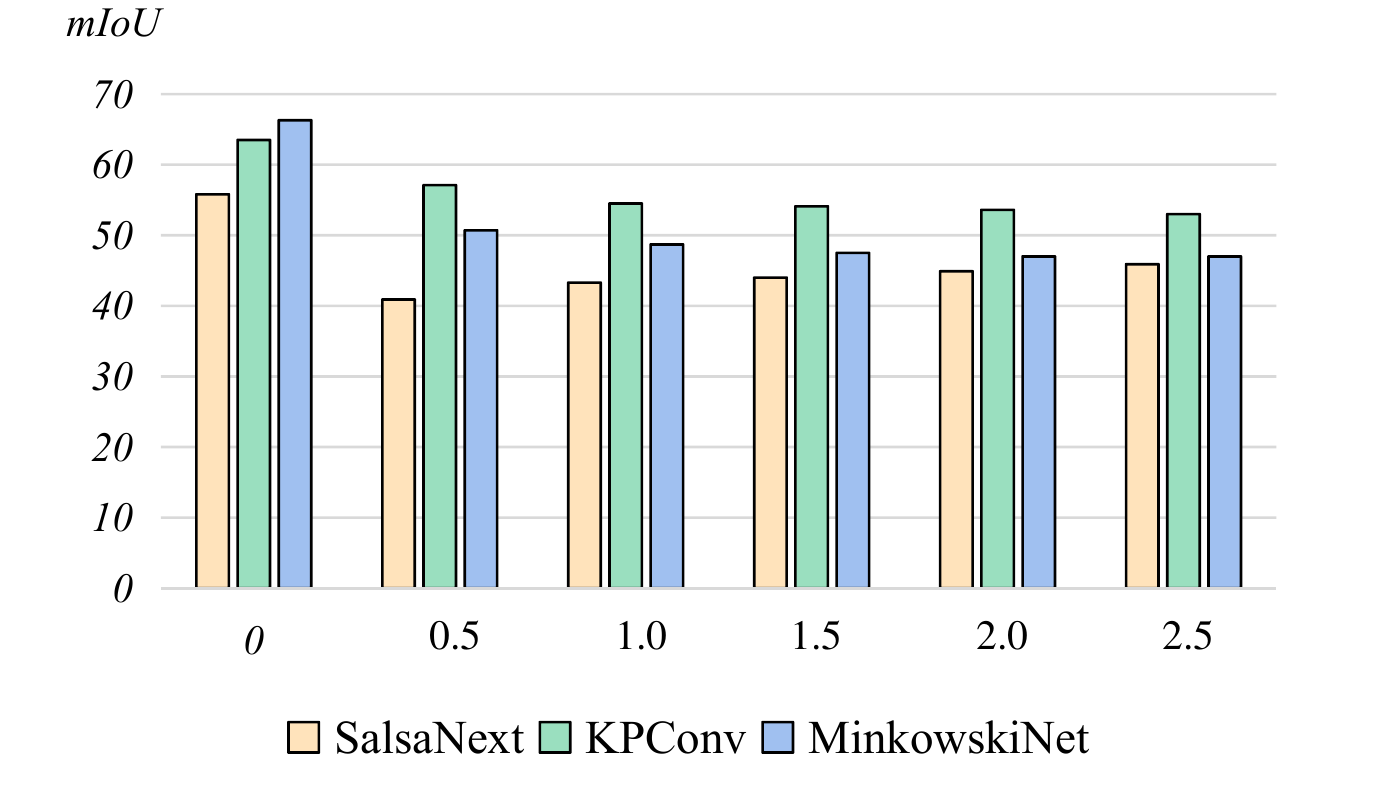}
		\caption{\textbf{Detailed results in snow simulation.} Performance of three typical approaches in different snow simulation intensities, where the x-axis denotes different snowfall rates (mm/h).}
		\label{fig:snow_results}
	\end{figure}

	\begin{table*}[t]
		\centering
		\caption{Comprehensive results on different LiDAR types. In `sparseness', we see the original LiDAR point as `dense' one, and randomly sample 1/2 points in each beam to generate a `sparse' one.}
		\begin{tabular}{cc|cc|cc|cc}
			\toprule
			&            & \multicolumn{2}{c|}{SalsaNext}         & \multicolumn{2}{c|}{KPConv}         & \multicolumn{2}{c}{MinkowskiNet}          \\
			LiDAR types  & Sparseness & mIoU      & R      & mIoU   & R      & mIoU         & R       \\\hline
			\multirow{2}{*}{64-beam} & Dense      & 55.8     & 100.0 & 63.0 & 100.0 & \textbf{66.3}        & 100.0  \\
			& Sparse     & 52.1     & 93.4  & 63.4  & 99.8  & \textbf{63.5}        & 95.8   \\\hline
			\multirow{2}{*}{32-beam} & Dense      & 52.4     & 93.9  & 59.0  & 92.9  & \textbf{62.5}       & 94.3   \\
			& Sparse     & 49.8     & 89.2  & 57.5  & 90.6  & \textbf{59.5}        & 89.7   \\\hline
			\multirow{2}{*}{16-beam} & Dense      & 32.3     &{57.8}  & 43.8  & 69.0  & \textbf{50.2}        & 75.7   \\
			& Sparse     & 29.7     & 53.2  & 43.0  & 67.7  & \textbf{46.4}        & 70.0  \\\bottomrule
		\end{tabular}
		\label{tab:cross-device}
	\end{table*}

	\noindent\textbf{Observation-3:} \textit{Transformer-based local aggregation greatly hampers the robustness, especially for global outliers.}
	
	Though transformer-based architecture improve the performance on the clean data, it greatly affects the robustness against diverse corruptions. Concretely, Point Transformer~\cite{zhao2021point} gains the lowest result in almost every corruption scenarios. 
	For instance, in the global outliers, most of the approaches can keep above 90\% performance, but it only achieves 61.5\% {on SemanticKITTI-C}.
	
	\noindent\textbf{Observation-4:} \textit{Pure voxel-based method shows most superior robustness cross all corruptions, especially for cross-devices scenario. Recent state-of-the-art voxel-based methods loss their robustness against corruption due to their hybrid-representation architectures.}
	
	The results illustrate that MinkowskiNet~\cite{choy20194d} enjoy most superior robustness with pure voxel architecture. Specifically, it achieves 76.0\% performance preserve {in the case of 16-beam cross-device on {SemanticKITTI-C,}} surpassing those of projection and point based methods over 20\% and 10\%, respectively.
	SPVCNN~\cite{tang2020searching} introduces point-wise MLP in parallel with voxel architecture,  nevertheless, it loses the robustness especially in fog simulation and 16-beam cross-device corruptions.
	Recent state-of-the-art Cylinder3D~\cite{zhu2021cylindrical} and 2DPASS~\cite{yan20222dpass} have poor generalization ability since they use extra representation. More analysis for this design will be illustrated in Sec.~\ref{sec:result3}.

	\subsection{Robustness in Specific Corruption}\label{sec:result2}
	In this section, we demonstrate and analyze the result of each specific corruption. To facilitate the experiment, we only select the most typical method in each mainstream, \ie, SalsaNext, KPConv and MinkowskiNet, {and test them on SemanticKITTI-C.}
	
	\noindent\textbf{Fog simulation.}
	Fig.~\ref{fig:fog_results} illustrates the comprehensive results of robustness in fog simulation. Apart from the three intensities related in Tab.~\ref{corruption_class}, \ie, 0.005, 0.06 and 0.2, we also provide other 6 intensities, including 0.01, 0.02, 0.03, 0.1, 0.12, 0.15.
	We find out that SalsaNext is greatly affected by denser fog, especially when $\beta$ is larger than 0.06. 
	Inversely, KPConv shows its superior robustness crossing different fog intensities.
	Therefore, we have the following summary:
	
	\noindent\textbf{Observation-5:} \textit{All types of methods will have performance decay as the fog becomes heavier, among which the projection-based method decreases fastest, and the point-based method decreases slowest.}

	\begin{figure}[t]
		\centering
		\includegraphics[width=\linewidth]{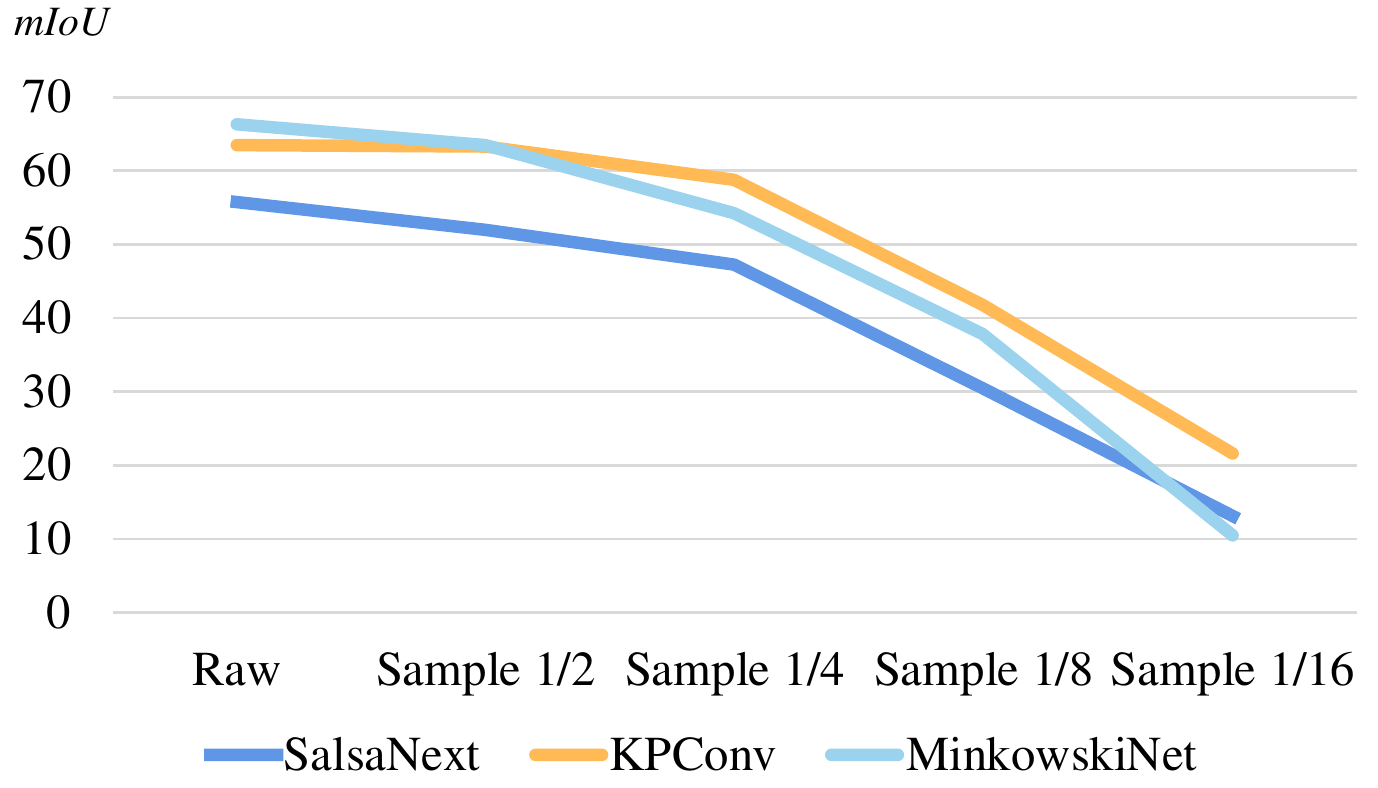}
		\caption{{\textbf{Results using different sample ratio per beam.}}}
		\label{fig:sample_results}
	\end{figure}
	
	\noindent\textbf{Snow simulation.}
	Fig.~\ref{fig:snow_results} illustrates the detailed results of snow simulation. The conclusion is:
	
	\noindent\textbf{Observation-6:} \textit{Though snow simulation hampers all types of methods, the performances of point-based and voxel-based methods are slightly decreased as the snow becomes heavier, while the projection-based methods illustrate an inverse tendency, \ie, there are slight performance boosts in heavier snowfall.}
	
\begin{figure}[t]
		\centering
		\includegraphics[width=\linewidth]{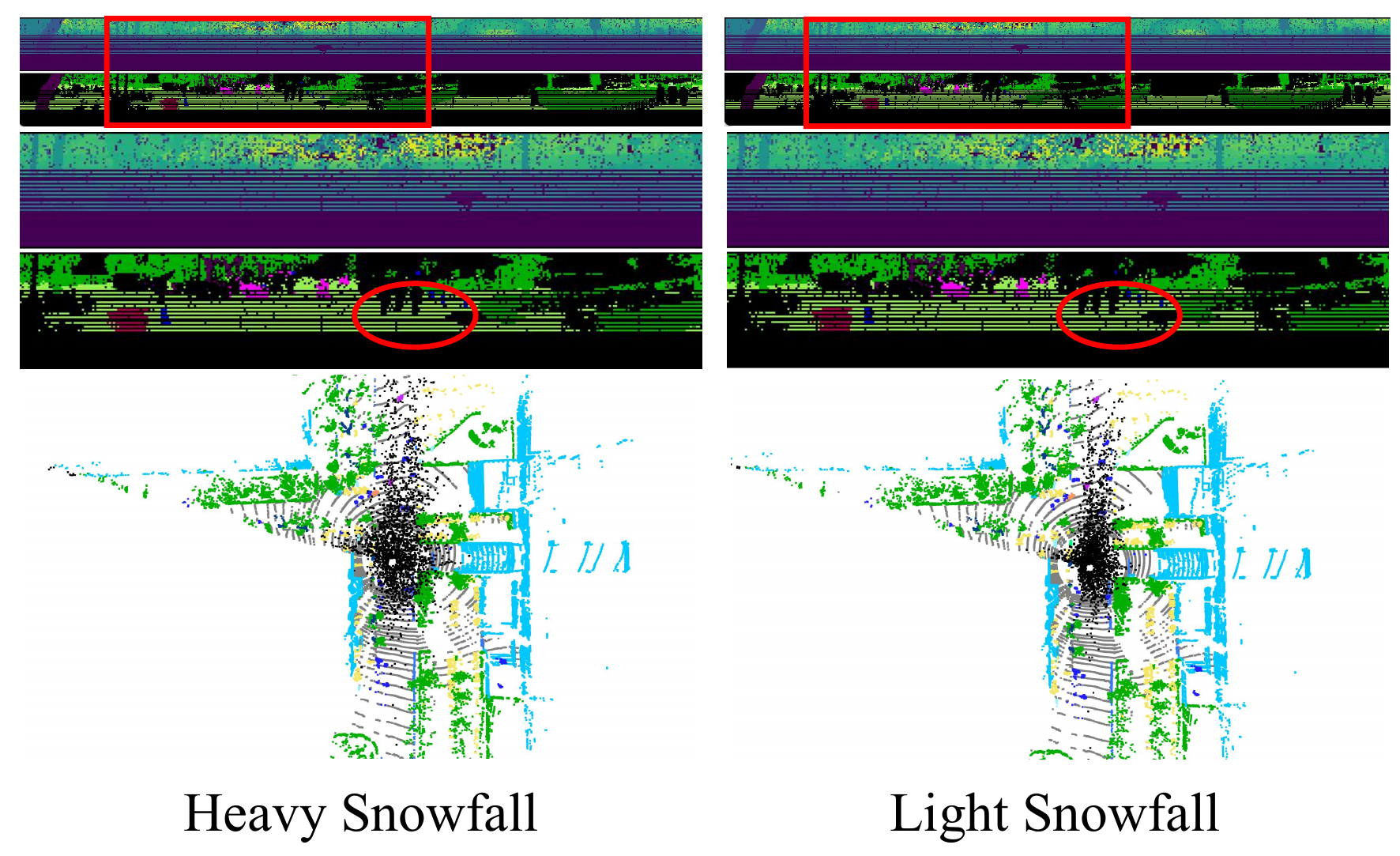}
		\caption{{\textbf{Range images in different snowfall intensities.} Although heavy snowfall occupies larger area in LiDAR point cloud, the noisy points will sparser in local areas, making less corruption on the range image.}}
		\label{fig:snow_range_robustness}
	\end{figure}

	The reason may be that the large snowfall makes the scattered noise in the 3D space cover a larger area, thus affecting the point-based and voxel-base methods.
	{However, as shown in Fig.~\ref{fig:snow_range_robustness}, when the 3D scene is mapped to a range image, these scattered points with larger coverage will become sparse in each pixel.}

	\noindent\textbf{Noisy corruptions.}
	Results in different noisy corruptions are illustrated in Tab.~\ref{tab:noisy}. On one hand,  MinkowskiNet is the most robust method against global outliers. It even keeps 95.8\% performance in the scenario with additional 50\% global noisy points. On the opposite, SalsaNext has poor generalization ability for global noise, especially with a larger proportion of noisy points.
	On the other hand, the projection-based method (\ie, SalsaNext) shows great robustness in local distortion noises, spanning different jittering ranges, as summarized in {observation-1}. In contrast, KPConv and MinkowskiNet cannot work normally in large-range jittering distortion.

	\noindent\textbf{Cross-device discrepancy.} We demonstrate concrete results on different LiDAR types in Tab.~\ref{tab:cross-device}, where MinkowskiNet achieves the best results in cross-device scenarios.
	In contrast, SalsaNext has poor robustness, especially in 16-beam devices with only around 50\% original performance.
	Furthermore, there is an interesting discovery: 
	
	\noindent\textbf{Observation-7:} \textit{Point-based methods (\ie, KPConv) are greatly robust against the scenario of downsampling points in each LiDAR beam.}

	As illustrated in the table, removing 1/2 points in each beam nearly does not affect the performance of KPConv.
	Specifically, in both 16 and 32-beam devices, the performance in sparse cases is greatly similar to the dense ones.
	{{Furthermore,} we evaluate the robustness of above three methods when facing point sample in each beam, as shown in Fig.~\ref{fig:sample_results}.
		We can find out that the point-based method shows stronger robustness in 1/16 sampled points, while the voxel-based method performs worst.}
	This achievement may come from the sampling process in point-based methods.
	Compared with projection-based and voxel-based methods that down-scale feature maps through pooling or convolution operations, point-based methods reduce the point numbers through sampling strategies, as depicted in Sec.~\ref{pointbase}. 
	Therefore, the local aggregations in point-based methods are more robust to downsampling operation.
	
	\begin{table*}
			\centering
			\caption{Systematic analysis for input representation, architecture design and data augmentation. The analysis includes (A) architecture for projection-based methods; (B) architecture for point-based methods; (C-D) representation for voxel-based methods; (E) hybrid-representation architecture and (F) data augmentation. The baseline models are marked with \underline{underline}. The increase and decrease compared with baselines are denoted as $\uparrow$ and $\downarrow$, respectively.}
			\resizebox{\textwidth}{!}{\begin{tabular}{cl|c|cc|cccc|cccc|cccc}
					\toprule
					&       & \textit{Clean} & \multicolumn{2}{c|}{\textit{Robustness}} & \multicolumn{2}{c}{\textit{Fog}} & \multicolumn{2}{c|}{\textit{Snowfall}} & \multicolumn{2}{c}{\textit{Global Outliers}} & \multicolumn{2}{c|}{\textit{Local Distortion}} & \multicolumn{2}{c}{\textit{32-beam}} & \multicolumn{2}{c}{\textit{16-beam}} \\
					Analysis	&   Method Descriptions   & {mIoU} & {RmIoU} & {mR} & {mIoU} & {R} & {mIoU} & {R} & {mIoU} & {R} & {mIoU} & {R} & {mIoU} & {R} & {mIoU} & {R} \\\hline
					\multirow{3}[0]{*}{{{A}}} & {\underline{CENet (2048$\times$ 64)}} & 64.3 & 49.3 & 76.6 & 32.5 & 50.5 & 53.0 & 82.4 & 59.0 & 91.7 & 60.4 & 94.0 & 56.3 & 87.5 & 34.5 & 53.7 \\
					& {CENet (1024$\times$ 64)} & 62.1 & ~~~47.5 $\downarrow$ & ~~~77.2 $\uparrow$ & ~~~32.0 $\downarrow$ & ~~~51.5 $\uparrow$ & ~~~48.3  $\downarrow$ & ~~~78.8 $\downarrow$ & ~~~56.5 $\downarrow$ & ~~~91.0 $\downarrow$ & ~~~56.9 $\downarrow$ & ~~~91.6 $\downarrow$ & ~~~57.8 $\uparrow$ &~~~93.0 $\uparrow$ & ~~~36.2 $\uparrow$ & ~~~58.2 $\uparrow$\\ 
					& {CENet (512$\times$ 64)} & 61.5 & ~~~43.3 $\downarrow$ & ~~~70.4 $\downarrow$ & ~~~30.5 $\downarrow$ &~~~49.6 $\downarrow$ &~~~42.3  $\downarrow$&~~~68.8 $\downarrow$ & ~~~53.7 $\downarrow$ & ~~~87.3 $\downarrow$ &~~~50.5 $\downarrow$&~~~82.0 $\downarrow$ & ~~~55.3 $\downarrow$ &~~~90.0 $\uparrow$& ~~~27.5 $\downarrow$ &~~~44.7 $\downarrow$ \\ 
					\hline
					\multirow{3}[0]{*}{{{B}}} & \underline{RandLA-Net} & 59.2 &47.6 & 80.4 & 56.4 & 95.4 &   50.0    &   84.4    & 57.8 & 97.8 & 26.4 & 44.7 &    53.3   &   90.1    &    41.7   & 70.5 \\
					& {w/o Attentive Pooling}  & 56.7 & ~~~43.5 $\downarrow$& ~~~76.8 $\downarrow$&~~~51.3 $\downarrow$ &~~~90.5 $\downarrow$&~~~47.3 $\downarrow$&~~~83.5 $\downarrow$&~~~55.4 $\downarrow$& ~~~97.7 $\downarrow$& ~~~26.9  $\uparrow$& ~~~47.5 $\uparrow$&~~~49.7 $\downarrow$&~~~87.6 $\downarrow$&~~~30.6 $\downarrow$ &~~~54.0 $\downarrow$\\
					& {Point-wise MLP} & 50.3 & ~~~33.5 $\downarrow$ &~~~66.7 $\downarrow$ & ~~~40.6 $\downarrow$ & ~~~80.7 $\downarrow$& ~~~33.4 $\downarrow$& ~~~66.5 $\downarrow$& ~~~48.8 $\downarrow$& ~~~97.1 $\downarrow$& ~~~11.8 $\downarrow$& ~~~23.5 $\downarrow$& ~~~39.0 $\downarrow$& ~~~77.6 $\downarrow$&~~~27.4 $\downarrow$& ~~~54.5 $\downarrow$\\
					\hline
					\multirow{2}[0]{*}{{{C}}} & \underline{MinkowskiNet (grid, 5cm)} & 66.3 & 53.6 & 80.9 & 56.3 & 84.9 & 50.4 & 76.1 & 65.3 & 98.5 & 37.0 & 55.9 & 62.2 & 93.9 & 50.4 & 76.0 \\
					& {MinkowskiNet (cylinder)} & 63.3 &~~~51.5 $\downarrow$ & ~~~81.3 $\uparrow$ & ~~~59.4 $\uparrow$ &~~~94.0 $\uparrow$&  ~~~58.9 $\uparrow$ & ~~~93.0 $\uparrow$&~~~62.7 $\downarrow$ & ~~~99.0 $\uparrow$ & ~~~45.3 $\uparrow$&~~~71.5 $\uparrow$& ~~~50.8 $\downarrow$& ~~~80.2 $\downarrow$& ~~~31.9 $\downarrow$& ~~~50.3 $\downarrow$\\\hline
					\multirow{2}[0]{*}{{{D}}} & {MinkowskiNet (10cm)} & 64.6 & ~~~50.7 $\downarrow$& ~~~78.6 $\downarrow$ & ~~~52.2 $\downarrow$& ~~~80.8 $\downarrow$ & ~~~46.7 $\downarrow$& ~~~72.3 $\downarrow$& ~~~57.0 $\downarrow$& ~~~88.2 $\downarrow$& ~~~38.0 $\uparrow$ & ~~~58.8 $\uparrow$& ~~~60.5 $\downarrow$ & ~~~93.6 $\downarrow$ & ~~~50.1 $\uparrow$& ~~~77.5 $\uparrow$\\
					& {MinkowskiNet (20cm)} & 60.6  &  ~~~43.7 $\downarrow$&~~~72.1 $\downarrow$& ~~~43.4 $\downarrow$&~~~71.6 $\downarrow$& ~~~37.3 $\downarrow$&~~~61.5 $\downarrow$& ~~~47.1 $\downarrow$&~~~77.7 $\downarrow$& ~~~32.5 $\downarrow$& ~~~53.7 $\downarrow$&  ~~~55.7 $\downarrow$& ~~~92.0 $\downarrow$& ~~~46.2 $\downarrow$& ~~~76.2 $\uparrow$ \\\hline
					\multirow{3}[0]{*}{{{E}}}& \underline{2DPASS} & 70.1 & 51.1 & 72.9 & 40.4 & 57.6 & 53.6 & 76.5 & 69.8 & 99.6 & 43.9 & 62.7 & 61.3 & 87.4 & 37.7 & 53.7\\
					& {2DPASS w/o PointHead}  & 65.4  & ~~~48.5 $\downarrow$ & ~~~74.2 $\uparrow$ &   ~~~44.2 $\uparrow$& ~~~67.7 $\uparrow$&~~~47.7 $\downarrow$&~~~72.9 $\downarrow$& ~~~65.0 $\downarrow$&~~~99.6 $\uparrow$& ~~~38.2 $\downarrow$& ~~~58.4 $\downarrow$& ~~~57.2 $\downarrow$&~~~87.5 $\uparrow$& ~~~38.7 $\uparrow$& ~~~59.1 $\uparrow$\\
					& {2DPASS w/o PointBranch} &64.1 & ~~~47.4 $\downarrow$ & ~~~73.9 $\uparrow$ & ~~~35.9 $\downarrow$ & ~~~56.0 $\downarrow$ &  ~~~46.9 $\downarrow$ & ~~~73.3 $\downarrow$ & ~~~63.6 $\downarrow$& ~~~99.3 $\downarrow$& ~~~45.0 $\uparrow$& ~~~70.2 $\uparrow$& ~~~56.3 $\downarrow$& ~~~87.9 $\uparrow$ & ~~~36.3 $\downarrow$ &  ~~~56.7 $\uparrow$\\\hline
					\multirow{2}[0]{*}{{{F}}} & {MinkowskiNet + InsCutMix} & 70.7  & ~~~58.5 $\uparrow$ &~~~82.7 $\uparrow$ &  ~~~58.7  $\uparrow$ & ~~~83.0 $\downarrow$ & ~~~59.6 $\downarrow$ & ~~~84.3 $\downarrow$ & ~~~70.0 $\downarrow$ & ~~~99.1 $\downarrow$ & ~~~48.3 $\downarrow$ &~~~68.3 $\downarrow$ & ~~~66.6 $\downarrow$ &~~~94.2 $\downarrow$ &~~~47.6  $\downarrow$ &~~~67.3  $\downarrow$\\
					& {MinkowskiNet + Mix3D} & 71.7 & ~~~59.5  $\uparrow$ & ~~~83.0  $\uparrow$ & ~~~57.4 $\uparrow$ & ~~~80.0 $\downarrow$& ~~~63.4 $\uparrow$ &~~~88.4  $\uparrow$& ~~~71.7 $\uparrow$ & ~~~100.0 $\uparrow$ & ~~~62.6 $\uparrow$ &~~~87.3 $\uparrow$ & ~~~63.6 $\uparrow$& ~~~88.7 $\downarrow$&~~~38.5  $\downarrow$&~~~53.7 $\downarrow$\\
					\bottomrule
			\end{tabular}}%
			\label{tab:design}%
	\end{table*}
	
	\subsection{Model Design v.s. Robustness}\label{sec:result3}
	In this section, we comprehensively analyze the relationships between different model designs and robustness. The results are illustrated in Tab.~\ref{tab:design}.
	
	\noindent\textbf{Size of range image (projection-based).}
	As shown in \textbf{Analysis~A} of the table, we train CENet~\cite{cheng2022cenet} with different sizes of range image, (\ie, $512\times 64$, $1024\times 64$ and $2048\times 64$), and gain the following conclusion:
	
	\noindent\textbf{Observation-8:} \textit{Exploiting smaller image size in range projection will make the model more vulnerable to noise, except for cross-device scenarios.} 
	
	When adopting $512\times 64$ range image as input, the performance of CENet decreases from 76.6\% R to 70.4\% R, especially in snowfall simulation with 14\% robustness drop.
	Moreover, its mIoU dramatically decays from 60.4\% to 50.5\% in LiDAR {data} with local noises.
	On the opposite, adopting smaller range images improves the robustness when deploying the model in devices with smaller beam numbers.
	We believe that the reason for the above phenomenon is that small images will make more points gather in the same pixel. Thus, noise points can easily enlarge the proportion of contaminated pixels. 
	However, in the LIDAR point cloud with a smaller beam number, a smaller image size makes the density of the valid pixels still quite high.
	
	\noindent\textbf{Local aggregation (point-based).}
	For the point-based approach, we select RandLA-Net as a typical one and conduct an ablation study, as shown in \textbf{Analysis~B} of Tab.~\ref{tab:design}.
	Specifically, we first replace the attentive pooling with max pooling in the second line, and replace the transformation function $\mathcal{T}$ (Eqn.~\eqref{eqn:randla}) to naive point-wise MLPs in the third line.
	The results show that both two components greatly improve the performance and the generalization ability.
	After discarding the two components, the ablated model can only achieve 50.3 mIoU on clean data and keep 66.7\% performance in the common corruptions.

	\noindent\textbf{Voxel partition (voxel-based).} 
	To further study the effectiveness of different voxel partitions, we conduct experiments and illustrate the results in \textbf{Analysis~C} of Tab.~\ref{tab:design}.
	During the experiments, we change the voxel partition of MinkowskiNet to the cylinder one and keep the network architecture the same.
	Concretely, we first transform the LiDAR point cloud from the Cartesian system into a polar coordinate system through Eqn.~\eqref{polar}. 
	After that, we discretize the transformed LiDAR data with voxel size [0.05, 0.001$\pi$, 0.05] in corresponding axes.
	Finally, the following summary can be obtained.

	\noindent\textbf{Observation-9:}  \textit{Cylindrical partition in voxelization greatly improves the robustness in most of the corruption, except in cross-device LiDAR data.} 
	
	Specifically, after exploiting cylindrical voxelization, the models' robustness in fog, snow and local corruptions are increased by around 10\%, 17\% and 16\%, respectively.
	However, such improvement is only for out-of-distribution data.
	The performance of the model for clean data is dropped to 60.5 mIoU (a drop of about 6\%). Similarly, its performance in the cross-device deployment scenario is also affected, especially the robustness on 16-beam LiDAR is reduced by 26\%.
	Nevertheless, robustness in different voxelization is still an important discovery in this paper, and it also lays a foundation to propose our newly configured method RLSeg in Sec.~\ref{sec:ours}.
	
	\noindent\textbf{Voxel size (voxel-based).}
	There are also experiments to study the robustness through different voxel sizes. The results are shown in  \textbf{Analysis~D} of Tab.~\ref{tab:design}.
	
	\noindent\textbf{Observation-10:}  \textit{Larger voxel partition makes voxel-based approaches more vulnerable to global-level corruptions, such as adverse weathers and global outliers. However, the robustness against local distortion and cross-devices point clouds is improved.} 
	
	In the experiment, we apply a larger voxel partition (\ie, 0.1m and 0.2m), compared with 0.05 as the origin.
	The results show that this setting greatly affects the robustness against global-level corruption, especially for the global noise. More importantly, utilizing a small voxel size makes the model cannot achieve satisfactory performance on clean data.
	The reason is that a large grid makes the model merge noisy and original points into the same grids, and loses fine-grained information.

	\begin{figure}[t]
		\centering
		\includegraphics[width=\linewidth]{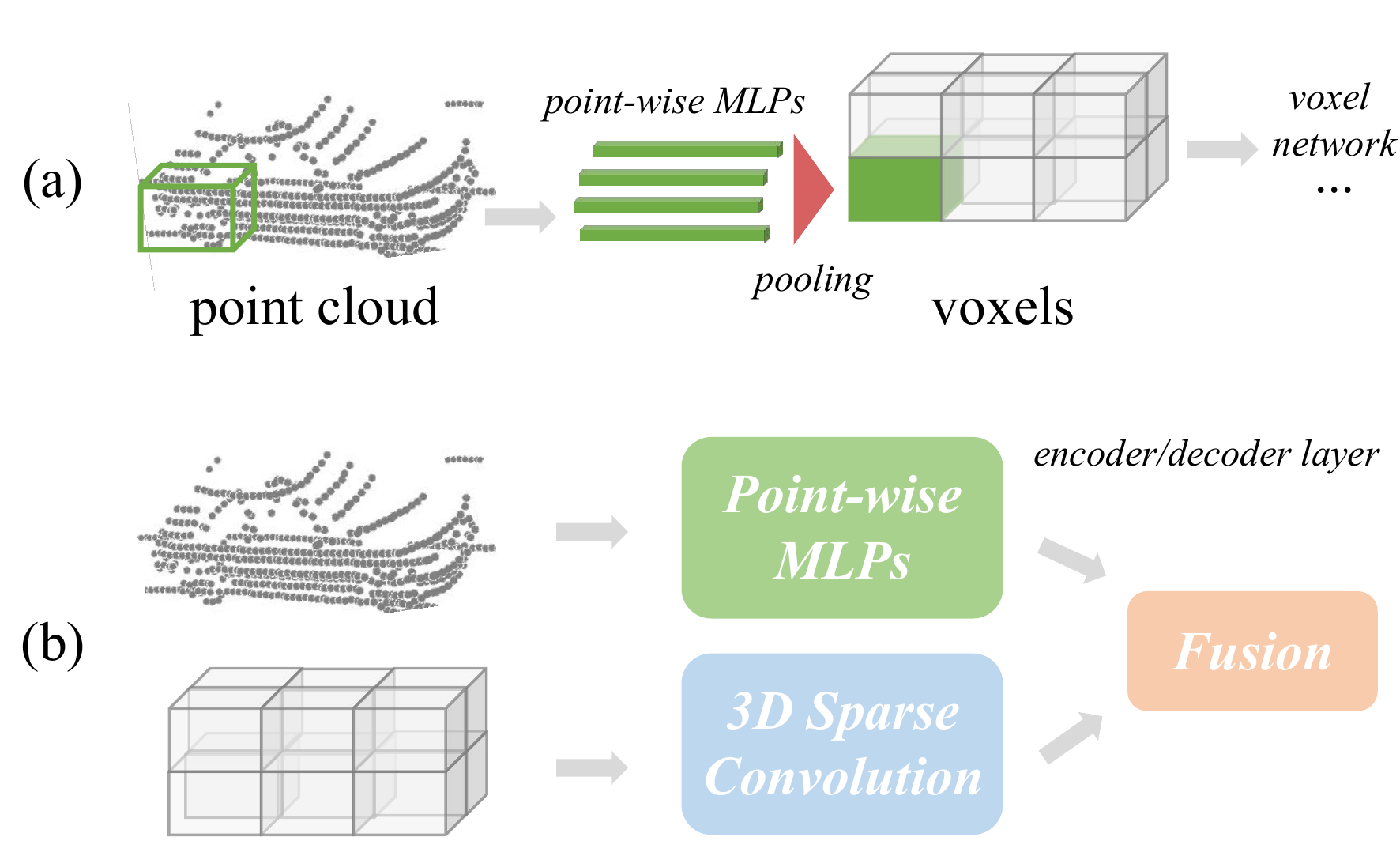}
		\caption{\textbf{Illustration of hybrid-representation architecture.} The architecture of PointHead and PointBranch are shown in (a) and (b).}
		\label{fig:multirep_robust}
	\end{figure}

	\noindent\textbf{Hybrid-representation architecture (voxel-based).}
	In  \textbf{Analysis~E} of Tab.~\ref{tab:design}, we investigate the relationship between robustness and hybrid-representation architectures.
	As mentioned in Sec.~\ref{voxelbase_revisit}, current state-of-the-arts adopt hybrid-representation architecture to boost the in-domain performance.
	Concretely, there are two components that merging point-wise representation into the voxel one, as shown in Fig.~\ref{fig:multirep_robust}.
	(a) PointHead: exploiting a PointNet architecture to aggregate point-wise features into individual voxel grids.
	(b) PointBranch: extracting point-wise features in parallel, and merging the features from voxel architecture.
	Similar components can be found in other previous works (\eg, Cylinder3D and SPVCNN), but here we only conduct ablation on the state-of-the-art.
	
	\noindent\textbf{Observation-11:}  \textit{Although hybrid-representation architectures improve the performance for the in-domain LiDAR segmentation with clean data, they are detrimental to model robustness, especially when using the PointHead component.} 
	
	As shown in Tab.~\ref{tab:design}, both PointHead and PointBranch boost the performance of 2DPASS on the clean LiDAR data.
	However, when the PointHead is exploited, there is a dramatic decrease in the robustness, especially in fog simulation and 16-beam device with 10\% and 5\% robustness drops.
	Similarly, PointBranch also hampers the robustness, but the influence is much slight.
	Directly conducting point-wise MLPs on LiDAR points is easier affected by diverse corruptions, since it cannot capture local geometric.

	\subsection{Data Augmentation v.s. Robustness}\label{sec:result4}
	In this section, we investigate the robustness of the model with different data augmentation.

	\begin{table*}[t]
		\centering
		\small
		\caption{Ablation study for RLSeg on SemanticKITTI-C. KD and PL denote knowledge distillation and pseudo label fine-tuning, respectively.}
		\resizebox{\textwidth}{!}{\begin{tabular}{l|ccc|c|cc|cc|cc|cc}
				\toprule
				Model   & Mix3D                     & KD                        & PL                        & mIoU & RmIoU & mR   & Fog  & Snowfall & Global & Local & 32-beam & 16-beam \\\hline
				\multirow{2}[0]{*}{MinkowskiNet} &  &                           &                           & 66.3 & 53.6  & 80.9 & 56.3 & 50.4     & 65.3   & 37.0  & 62.2    & \textbf{50.4 }   \\
				& \checkmark &                           &                           & 71.7 & 59.5  & 83.0 & 57.4 & 63.4     & 71.7   & 62.6  & \textbf{63.6}    & 38.5    \\\hline
				\multirow{4}[0]{*}{RLSeg (ours)}       &                           &                           &        &   63.3     &   51.5    &   81.3   &   {59.5}  &    58.9      &    62.7    &   45.3    &    50.8    &    31.9     \\
				& \checkmark &                           &                           &    67.0  &  56.7      &    84.7  & \textbf{60.4}     & 63.5         &  66.3      &   53.2    &      58.2   &    38.7     \\
				& \checkmark & \checkmark &                           &    70.9  &  60.3     &  85.0    &   55.9  &    64.0     &   70.8    &   69.6   &    60.0     &    41.5     \\
				& \checkmark & \checkmark & \checkmark &    \textbf{73.5}  &   \textbf{62.5}    &  \textbf{85.0}    &   {57.6}   &    \textbf{66.2}     &    \textbf{73.4}  &   \textbf{71.9}    &   {62.3}     &  {43.6}  \\
				\bottomrule    
		\end{tabular}}
		\label{tab:ablation}
	\end{table*}

	\noindent\textbf{General augmentation strategies.}
	Previous studies adopt diverse data augmentation during the training. Generally, rotation and scaling are the most widely used. In this paper, we conduct rotation, scaling, and flipping when re-training the point-based and voxel-based models, and follow the same image-based augmentation in projection-based models.
	Note that \textbf{jitter augmentation is not used in our experiment}, as it will generate in-domain training data for our local distortion corruption.
	
	\begin{figure}[t]
		\centering
		\includegraphics[width=\linewidth]{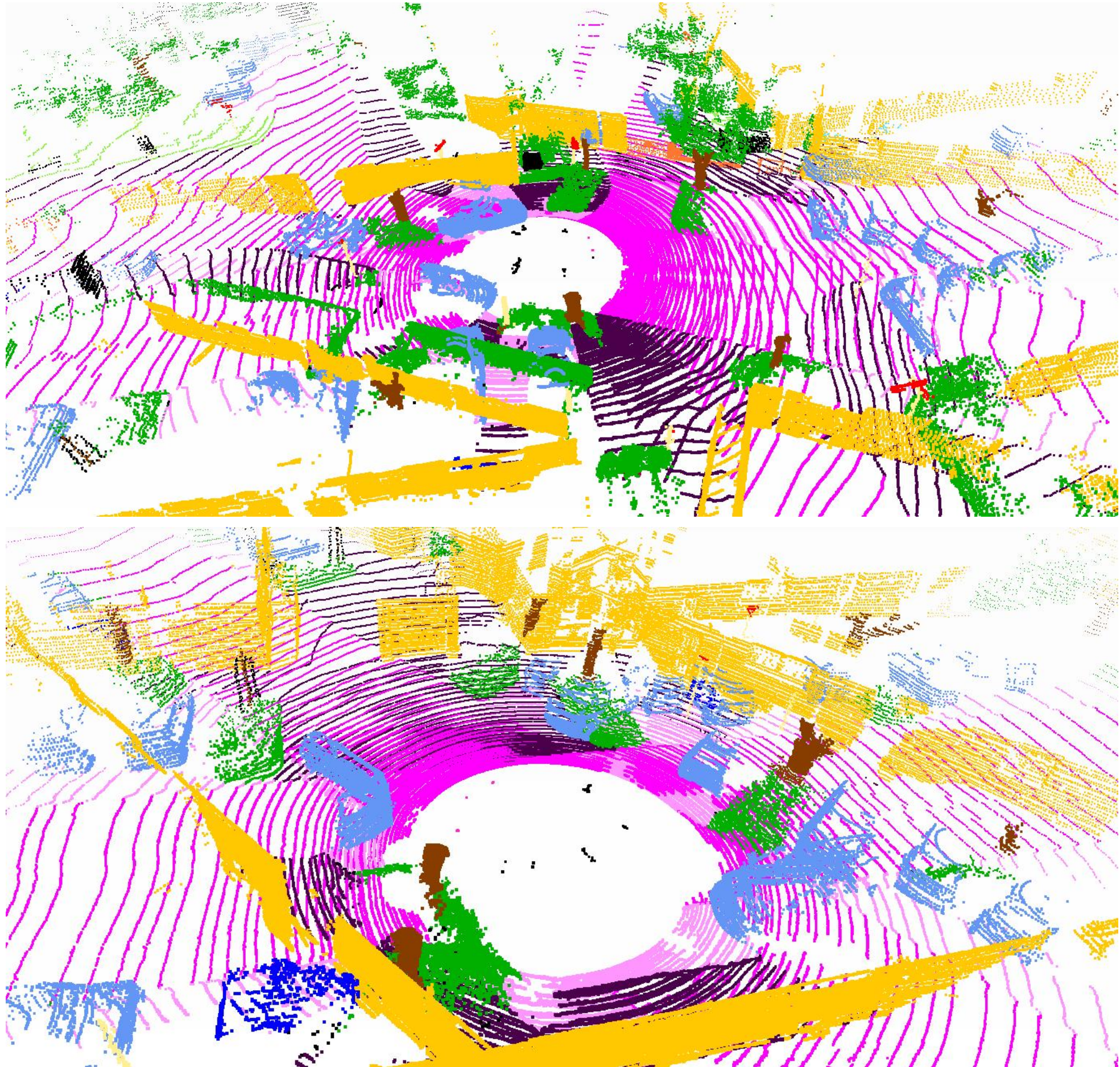}\vspace{.3cm}
		\caption{\textbf{Examples of LiDAR point cloud after applying Mix3D.}  }
		\label{fig:mixup}
	\end{figure}
	
	\begin{figure*}[t]
		\centering
		\includegraphics[width=\linewidth]{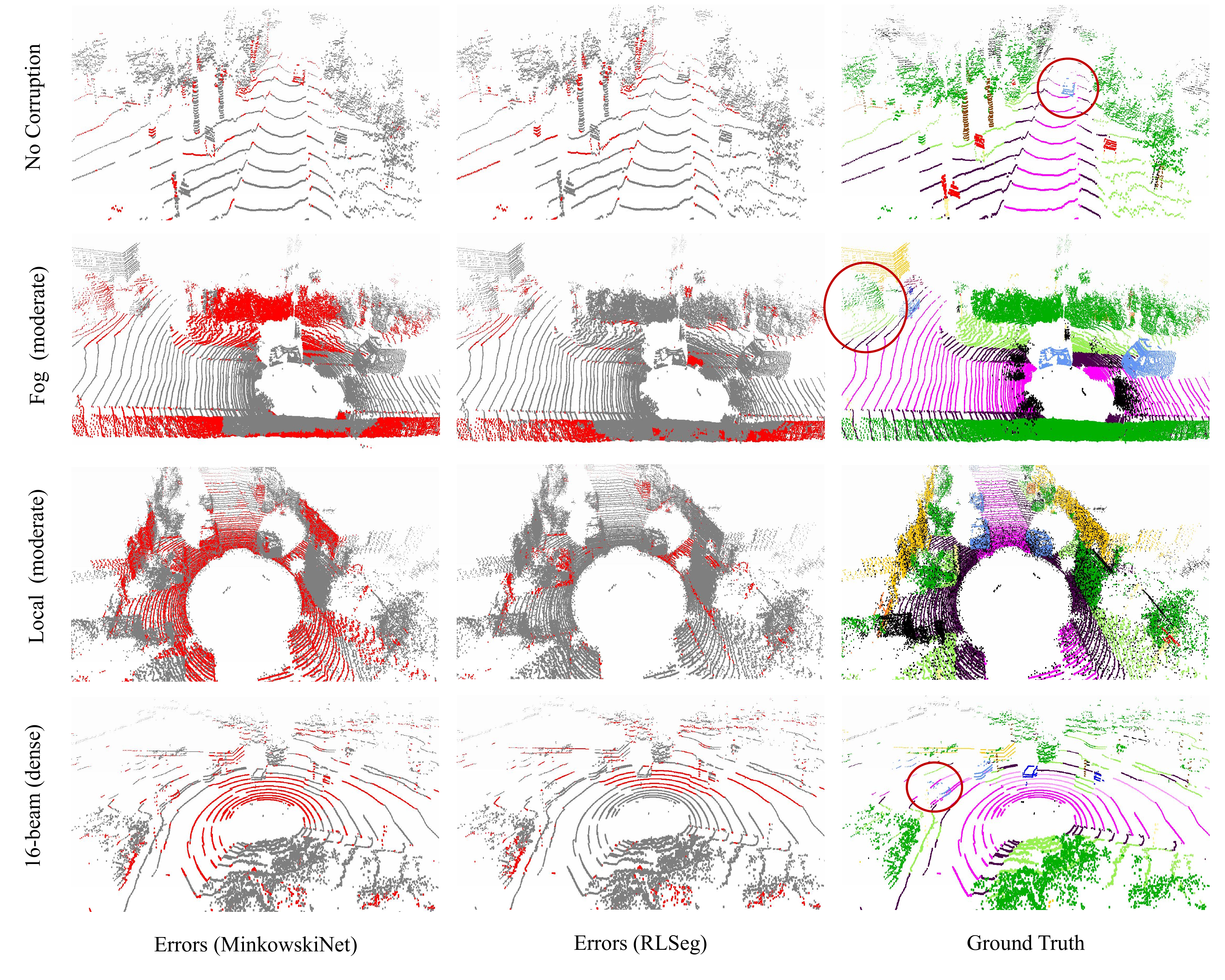}
		\caption{\textbf{Visualization.} We demonstrate the visualization results of the most robust existing method (MinkowskiNet) and our RLSeg on four cases, including clean data and three LiDAR corruptions. The noisy points are labeled as `ignore' (black color) and not considered in the evaluation. The left two columns are colorized by error maps, and the last one is colorized by ground truth. }
		\label{fig:visualization}
	\end{figure*}
	
	\noindent\textbf{MixUp on LiDAR point cloud.} MixUp~\cite{zhang2017mixup} is initially proposed in image classification for a more robust representation and extends to 3D computer vision in recent years. Existing MixUp approaches in LiDAR semantic segmentation task include Mix3D~\cite{Nekrasov213DV}, Instance CutMix~\cite{xu2021rpvnet} and LaserMix~\cite{kong2022lasermix}, where only Mix3D is open-sourced now.
	Therefore, we train MinkowskiNet with the official Mix3D, as well as the re-produced Instance CutMix.
	The illustration of Mix3D is in Fig.~\ref{fig:mixup}, in which Mix3D randomly merges two LiDAR scans (including labels) into a common coordinate.
	We re-produce Instance CutMix by only merging instance-level objects from other LiDAR scans.
	The experimental results are demonstrated in \textbf{Analysis~F} of Tab.~\ref{tab:design}.

	\noindent\textbf{Observation-12:}  \textit{Existing MixUp data augmentation for LiDAR semantic segmentation makes the model more robust against most of the corruption, except for cross-device scenarios.} 
	
	As illustrated in the table, after exploiting two MixUp augmentation, there are significant boosts in several corruptions, \eg, over 10\% and 30\% improvements on snowfall and local distortion corruptions with Mix3D augmentation.
	However, the robustness of the model decreases in cross-device LiDAR data, especially in 16-beam data with larger domain discrepancies.
	The reason is that MixUp augmentation utilizes denser mixed point clouds as input, and thus makes the model vulnerable to the sparse point clouds.

	\begin{figure}
		\centering
		\includegraphics[width=\linewidth]{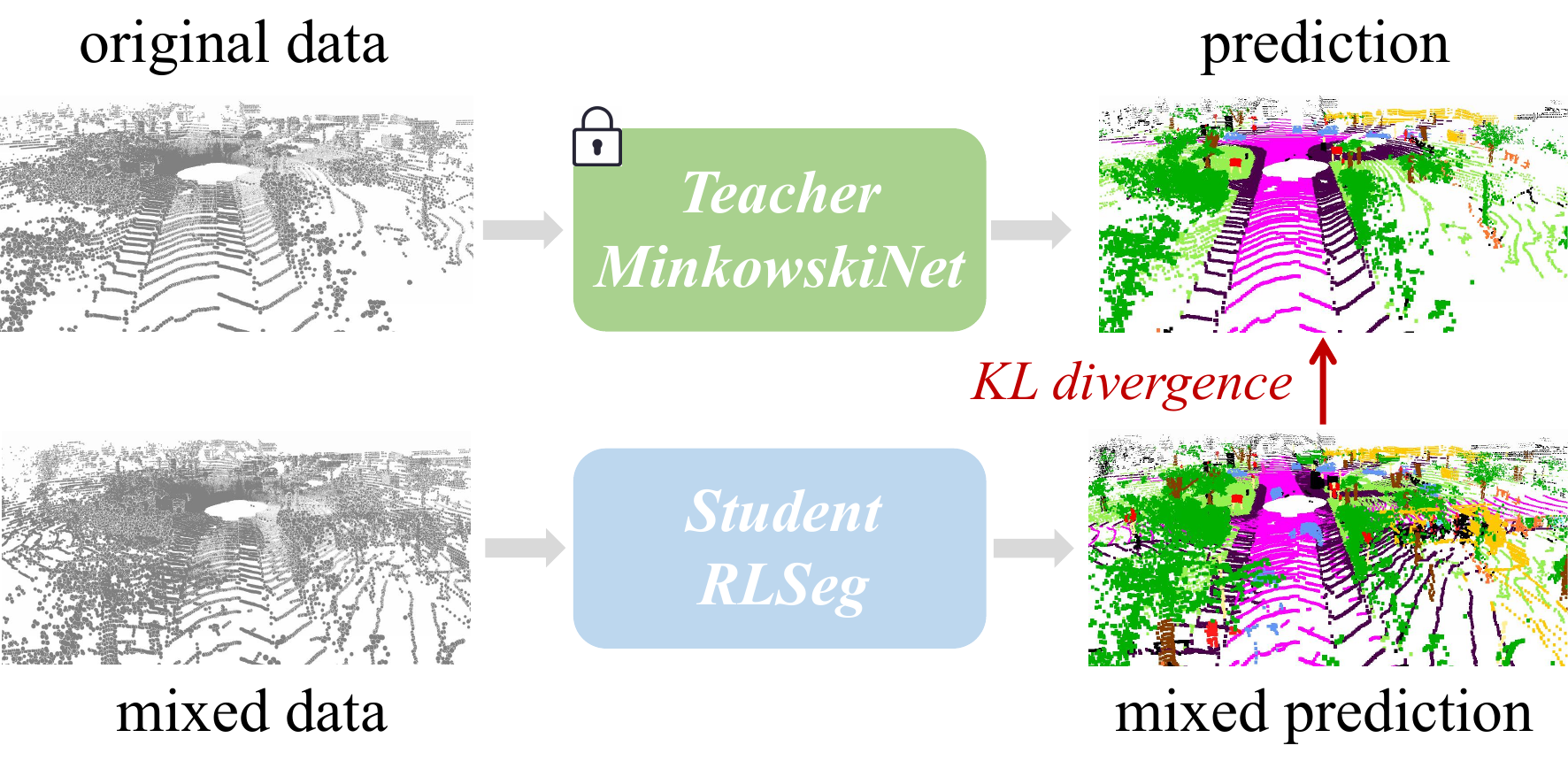}
		\caption{{\textbf{Illustration of training process of RLSeg.} We first train a teacher MinkowskiNet model with grid voxel partition and Mix3D augmentation.
				After that, RLSeg is trained with a teacher-student framework and apply KL divergence to the output logits of RLS and the teacher model. 
				During the training, the Mix3D is only conducted on the student model.}}
		\label{fig:rlseg}
	\end{figure}

	\section{{Boosting Corruption Robustness}}\label{sec:ours}
	
	\subsection{Summarize Observations}  
	In Sec.~\ref{sec:result}-\ref{sec:result4}, we obtain  12 observations in total.
	In order to design a more robust model, we summarize the most useful information from these observations:
	
	\noindent\textbf{1) Voxel-based architecture:} After summarizing observations 1-6, we prefer to use the voxel-based method as our backbone. The reason is that the methods of projection-based and point-based do not perform well on clean data, and both of them are vulnerable to certain corruption. In contrast, the voxel-based method performs well in most cases, and the results on clean data are also satisfactory.
	
	\noindent\textbf{2) Cylindrical partition with appropriate voxel size:} As shown in observation-9 and 10, exploiting cylindrical partition and appropriate voxel size increase the robustness of the model.
	
	\noindent\textbf{3) Single-representation:} Observation-11 illustrates that though hybrid-representation architecture improves the performance on clean data, it hampers the robustness against common corruptions.
	
	\noindent\textbf{4) Mix3D augmentation:} This can be gained by observation-12.

	\subsection{Robust LiDAR Segmentation} 
	{\noindent\textbf{Architecture.}
		Based on the above four conclusions, we design the robust LiDAR segmentation (RLSeg) model in this paper. 
		Specifically, we use MinkowskiNet as our backbone due to its single-representation nature.
		Furthermore, we transform the LiDAR point cloud from the Cartesian system into a polar coordinate system through Eqn.~\eqref{polar} and apply a cylindrical partition with voxel size [0.05, 0.001$\pi$, 0.05].
		However, the result in Tab.~\ref{tab:design} shows that cylindrical partition hampers the performance on the clean data.}
	
	\noindent{\textbf{Training through distillation.} To tackle this problem, we adopt the knowledge distillation~\cite{hinton2015distilling} to enhance the model.
		Specifically, we first train a teacher MinkowskiNet model with grid voxel partition, voxel size 0.05 and Mix3D augmentation, obtaining the teacher model, as shown in the last row of Tab.~\ref{tab:design}.
		After that, we train the above RLSeg with a {teacher-student framework}, applying KL divergence to the output logits of RLS and the teacher model, as shown in Fig.~\ref{fig:rlseg}. 
		During the training, the Mix3D is only conducted on the student model, and the KL divergence constrains the original data.
		We train RLSeg with 64 epochs with a weight of 0.05 for KL divergence.}
	
	\noindent{\noindent\textbf{Self-training with pseudo labels.} Motivated by the improvement achieved by pseudo label~\cite{sdseg3d_eccv2022} in semi-supervised learning, we further fine-tune the student network with 48 epochs on clean validation data with pseudo labels generated by the teacher MinkowskiNet.
		It can further improve the performance since the pseudo labels is generated by test-time augmentation.
		After the self-training process, the student model gains higher performance through conducting test-time augmentation again.}
	
	\subsection{Concrete Results} 
	{\noindent\textbf{Robustness evaluation.} Through such a simple but effective manner, we significantly improve the performance on clean data, while keeping the robustness against diverse corruptions.
		The results are demonstrated in Tab.~\ref{tab:benchmark}, where RLSeg significantly outperforms existing methods.
		Specifically, RLSeg can get the best performance on almost all corruption, especially global outliers and local distance, which reach 99.9\% and 97.8\% of the original performance respectively.
		At the same time, it can achieve 90.1\% original performance on snowfall corruption, which is 16.2 mIoU higher than its teacher model.
		Tab.~\ref{tab:benchmark2} further demonstrates RLSeg achieves state-of-the-art robustness on SemanticPOSS-C benchmark.
		Although the performance of RLSeg on clean SemanticPOSS is 0.9 mIoU less than that of SPVCNN, it keeps 84.2\% original performance while SPVCNN only gains 75.7\%.
		Note that SemanticPOSS is a smaller dataset compared with SemanticKITTI, and thus models trained on this dataset may generally have less robustness.
		Nevertheless, RLSeg still achieves comparable robustness as the model trained on SemanticKITTI.}
	
	\noindent{\noindent\textbf{Visualization.} We provide visualization results of our RLSeg and teacher MinkowskiNet in Fig.~\ref{fig:visualization}, in which our proposed model performs better.
		Specifically, MinkowskiNet cannot work normally in fog simulation and local distortion, and there are large areas of errors in the LiDAR scenes.
		In contrast, our RLSeg provides robust prediction even for the small objects (as shown in red circles).
		These show the robustness of our model as well as the promising future for robust LIDAR semantic segmentation.}
	
	\noindent{\noindent\textbf{Ablation study.} We analyze different designs through an ablation study on SemanticKITTI-C.
		As shown in Tab.~\ref{tab:ablation}, exploiting our architecture improves the robustness to 81.3\%, but causes a performance drop to 63.3 mIoU.
		After using Mix3D, there is a huge performance boost from 63.3 to 67.0 mIoU, while increasing the robustness from 81.3\% to 84.7\%.
		Although knowledge distillation (KD) does not improve the robustness significantly, it boost the performance on clean data from 67.0 to 70.9.
		Finally, utilizing pseudo label fine-tuning can further improve the performance while keeping the robustness, which shows a promising improvement by leveraging the potential of semi-supervised learning, giving a performance boost to about 73.5\%.
		This also gives hints for future work that improve the robust but poor performance network to better performance.}

	\section{Conclusion}\label{sec:conclusion}
	In this paper, {we propose new benchmarks called {SemanticKITTI-C} and SemanticPOSS-C, with respect to real-world and out-of-domain LiDAR corruptions.}
	We systematically investigate a wide range of LiDAR semantic segmentation models, spanning different input representations and network architectures.
	After analyzing the results of previous approaches, we summarized 12 observations for the future research.
	Finally, we propose RLSeg based on the above observations, which effectively boosts the robustness of LiDAR semantic segmentation. 
	We hope our benchmark, comprehensive analysis and observations could boost future research for robust LiDAR semantic segmentation in safety-critical applications.

	\backmatter
	
	%
	%
	%
	%
	
	
	
	\section*{Declarations}
	\bmhead{Supplementary information}
	Our dataset and code will be available at \url{https://yanx27.github.io/RobustLidarSeg/}.
	
	\bmhead{Funding}
	
	This work was supported in part by the Basic Research Project No. HZQB-KCZYZ-2021067 of Hetao Shenzhen HK S\&T Cooperation Zone, by the National Key R\&D Program of China with grant No.2018YFB1800800, by Shenzhen Outstanding Talents Training Fund, by Guangdong Research Project No. 2017ZT07X152 and No. 2019CX01X104, by the Guangdong Provincial Key Laboratory of Future Networks of Intelligence (Grant No. 2022B1212010001), by the NSFC 61931024\&8192 2046, by NSFC-Youth 62106154\&62302399, by zelixir biotechnology company Fund, by Tencent Open Fund, and by ITSO at CUHKSZ.
	
	\bmhead{Competing interests}
	The authors have no competing interests to declare that are relevant to the content of this article.
	
	\bmhead{Availability of data and materials}
	All the datasets used in the paper are publicly available.

	\bmhead{Authors' contributions}
	X. Yan and D. Dai conceptualized the
	work and designed the methodology. X. Yan and C. Zheng formulated the mathematical formulation. X. Yan conducted the experiments on SemanticKITTI-C {and Y. Xue conducted the experiments on SemanticPOSS-C.} X. Yan, D. Dai, and Z. Li analyzed the results. D. Dai, Z. Li and S. Cui supervised the work. All authors wrote and revised the manuscript.

	\bibliography{main.bbl}
	

\end{document}